\documentclass[a4paper,fleqn]{cas-sc}

\usepackage[numbers]{natbib}
\usepackage{amsmath}
\usepackage{booktabs}
\usepackage{multirow}

\renewcommand{\ttdefault}{lmtt}

\begin{document}
\let\WriteBookmarks\relax
\def\floatpagepagefraction{1}
\def\textpagefraction{.001}

\shorttitle{FRWKV+: Periodic-Aware Adaptive Gating}
\shortauthors{Yang et al.}

\title[mode=title]{FRWKV+: Periodic-Aware Adaptive Gating for Frequency-Space Linear Time Series Forecasting}

\author[1]{Qingyuan Yang}[orcid=0009-0005-6662-3141]
\ead{yqyneu@foxmail.com}

\author[1,2,3]{Dongyue Chen}
\ead{chendongyue@ise.neu.edu.cn}

\author[1]{Da Teng}[orcid=0009-0006-7996-3716]
\ead{13166672732@163.com}

\author[1]{Junhua Xiao}[orcid=0009-0007-8406-3461]
\ead{xiaojh1@mails.neu.edu.cn}

\author[1]{Jiaji Pan}[orcid=0009-0001-1307-7870]
\ead{1941310259@qq.com}

\author[1,2]{Shizhuo Deng}[orcid=0000-0002-6863-8516]
\cormark[1]
\ead{dengshizhuo@mail.neu.edu.cn}

\affiliation[1]{organization={College of Information Science and Engineering, Northeastern University},city={Shenyang},postcode={110819},state={Liaoning},country={China}}
\affiliation[2]{organization={Foshan Graduate School of Innovation, Northeastern University},city={Foshan},postcode={528311},state={Guangdong},country={China}}
\affiliation[3]{organization={National Frontiers Science Center for Industrial Intelligence and Systems Optimization},city={Shenyang},postcode={110819},state={Liaoning},country={China}}
\cortext[cor1]{Corresponding author.}

\begin{abstract}
Accurate and efficient long-term multivariate time series forecasting requires capturing recurring temporal structure while keeping inference cheap across many variables and horizons. Frequency-space models represent long-range and periodic variation compactly, but they typically process the real and imaginary spectral components as weakly coupled streams and treat periodic cues as ordinary input features, even when such cues are unreliable. This paper proposes FRWKV-Plus, a lightweight periodic-aware frequency-space forecasting model built on the efficient FRWKV backbone. FRWKV-Plus introduces a cross-branch spectral gate that reweights each spectral branch using a summary of its sibling branch, and a trust-gated residual correction that converts compact within-period context into a bounded, sign-flexible adjustment of these gates under a learned, data-dependent trust score. By construction, the correction is identity-preserving at initialization and strictly bounded, so periodic evidence can refine but never dominate or invert the base interaction. On seven standard benchmarks, FRWKV-Plus is consistently competitive with strong linear, frequency-domain, recurrent-style, and Transformer-based forecasters while preserving the lightweight profile of the backbone. Controlled three-seed ablations show that each component contributes, that the benefit is modest on strongly periodic data and pronounced on the harder Exchange and ILI datasets, and that the within-period context is the most influential single component. The implementation is publicly available at \url{https://github.com/yangqingyuan-byte/FRWKV-plus}.
\end{abstract}


\maketitle

\section{Introduction}
Long-term multivariate time series forecasting is an essential task in data-driven decision systems. It supports energy-management applications such as load and renewable-power forecasting \cite{gao2023adaptive}, transportation planning through spatiotemporal traffic prediction \cite{ji2023spatiotemporal}, and financial risk analysis through market forecasting \cite{cheng2022financial}. It is also central to weather monitoring \cite{wu2023interpretableweather} and healthcare surveillance \cite{morid2023healthcare}. In these scenarios, future observations are shaped by multiple forms of temporal structure, including local fluctuations, recurring periodic patterns, cross-variable dependencies, and long-range trends. A practical forecasting model must therefore satisfy two requirements simultaneously: it should represent complex temporal regularities with sufficient accuracy, while remaining efficient for repeated prediction across many variables and horizons.

A wide range of forecasting methods have been developed to meet these requirements. Recent surveys describe the field as moving from classical statistical tools toward neural sequence models, attention-based forecasters, decomposition-aware designs, frequency-space representations, and large pretrained forecasting systems \cite{wen2023transformers}. These directions have advanced the accuracy-efficiency trade-off, but much of the progress is organized around either improving temporal representation or reducing computational cost. Less attention has been paid to a narrower but important question: how recurring temporal evidence should regulate the internal interaction of frequency-space components.

Frequency-space forecasting provides a natural perspective for this question. Spectral transformation represents long-range and periodic variations compactly, making it attractive for efficient long-horizon prediction. However, when a real-valued time series is transformed into the complex spectrum, its temporal structure is no longer represented by a single real-valued feature stream. Instead, it is encoded by coupled real and imaginary components, which jointly determine magnitude- and phase-related information. Treating these components as largely separate streams can preserve computational simplicity, but it may also weaken the exchange of complementary information between them. This limitation is particularly relevant for lightweight frequency-space models, where explicit interaction modules must be carefully designed so that they improve representation without undermining efficiency.

Periodic-position information offers a useful conditioning signal for such interaction. A sample's position within a recurring cycle can indicate how current observations align with repeated temporal patterns. In the frequency domain, temporal alignment is closely related to phase-sensitive behavior, because changes in temporal position are reflected not only in spectral magnitudes but also in the relationship between real and imaginary components. Therefore, periodic-position context is not merely an auxiliary time feature. It can provide evidence about when and how the two spectral branches should communicate. This motivates a more targeted use of periodic information: instead of adding it directly to the representation, we use it to regulate real-imaginary branch interaction.

Recent large pretrained and cross-modality forecasting systems provide another route to broad temporal representation, often by drawing on language-like sequence modeling or auxiliary semantic cues. These systems broaden forecasting capability, but they also highlight a complementary need for compact architectures whose use of periodic evidence is explicit, selective, and computationally light.

At the same time, periodic cues should not be treated as universally reliable. The usefulness of within-period information may vary across samples, variables, datasets, and prediction horizons. If periodic features are injected as ordinary representation components, the model must absorb them even when their relevance to the current frequency state is weak. A more robust strategy is to use periodic-position information as conditional correction evidence. Under this view, the main frequency representation is preserved, while periodic context provides a controlled adjustment to the interaction pathway. The adjustment should be able to strengthen or weaken cross-branch exchange, and its influence should depend on the compatibility between periodic evidence and the current spectral state.

Based on this motivation, we propose \textit{FRWKV-Plus}, an efficient periodic-aware frequency-space forecasting model. FRWKV-Plus is built on the FRWKV backbone, which adopts lightweight RWKV-style frequency encoding for long-term time series forecasting \cite{yang2025frwkv}. On top of this backbone, FRWKV-Plus introduces cross-branch spectral gates (CSG) that reweight each spectral branch using a frequency-averaged summary of its sibling branch. It further develops a Periodic Positional Context Encoder (PPCE) to summarize compact within-period context, and uses a Trust-Gated Periodic Correction (TGPC) module to convert this context into signed residual corrections for the cross-branch gates. The correction is regulated by adaptive trust scores learned from the real branch, the imaginary branch, and the periodic-position context, so the model can admit periodic evidence selectively rather than relying on it unconditionally.

This design follows a simple principle: periodic information should refine frequency-space interaction instead of overwriting frequency-space representation. The base cross-branch gates provide a stable interaction path between real and imaginary components. The signed correction then acts as a bounded residual adjustment, allowing periodic context to increase or decrease the strength of branch communication. Because the correction is modulated by trust scores, FRWKV-Plus preserves the lightweight behavior of the original frequency-space backbone while introducing a mechanism that is sensitive to recurring temporal structure. In this way, the model connects within-period temporal context with cross-branch spectral interaction through an explicit and efficient gating pathway.

We evaluate FRWKV-Plus on seven widely used long-term forecasting benchmarks covering ETT, Weather, Exchange, and ILI datasets. The experimental results show that FRWKV-Plus is consistently competitive with representative forecasting baselines, including lightweight linear models, frequency-domain models, recurrent-style models, and Transformer-based forecasters. Module analyses further verify the effectiveness of periodic-position context, signed gate correction, and adaptive trust control. Efficiency profiling also shows that the proposed design maintains the lightweight computational profile of the frequency-space backbone, supporting its use as an accuracy-efficiency aware forecasting model.

The main contributions of this paper are summarized as follows.
\begin{itemize}
    \item We propose \textit{FRWKV-Plus}, a lightweight periodic-aware frequency-space forecasting model that augments the FRWKV backbone with cross-branch spectral gating and a periodic-position-conditioned correction, without adding a separate heavy representation branch.
    \item We introduce Trust-Gated Periodic Correction, a module that converts compact within-period context into a bounded, sign-flexible adjustment of the cross-branch spectral gates, admitted under a learned data-dependent trust score. The correction is identity-preserving at initialization and strictly bounded, so periodic evidence can refine but never dominate or invert the base interaction.
    \item Through controlled three-seed ablations, we show that each component contributes, that the within-period context is the most influential single component, and that the benefit of the design concentrates on non-stationary, data-scarce regimes (Exchange, ILI) rather than on strongly periodic data.
    \item We evaluate FRWKV-Plus on seven forecasting benchmarks together with module ablations and efficiency profiling, showing that it remains competitive with strong linear, frequency-domain, recurrent-style, and Transformer-based forecasters while preserving the lightweight profile of the backbone.
\end{itemize}

\begin{center}
\begin{minipage}{\textwidth}
\centering
\includegraphics[width=0.30\textwidth]{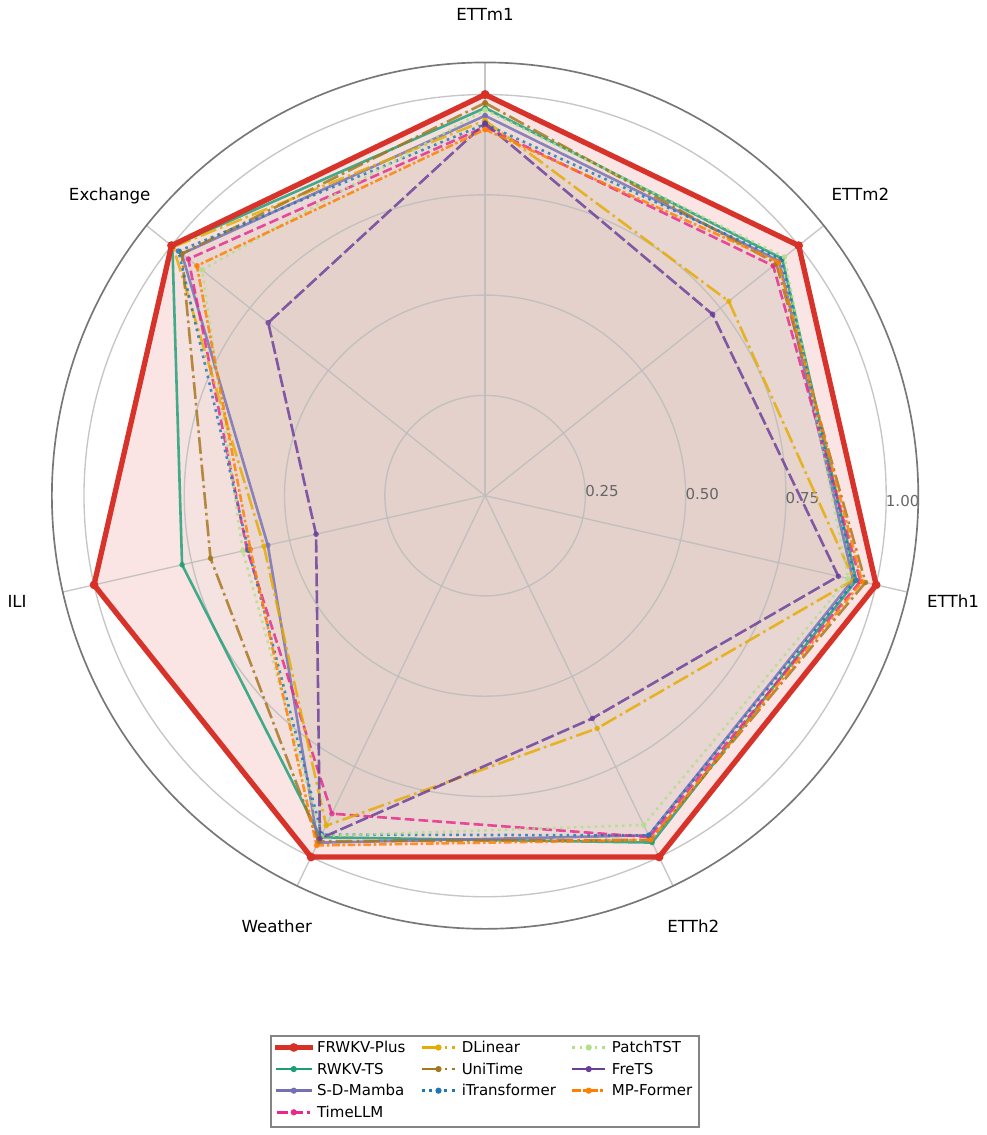}
\hspace{0.06\textwidth}
\includegraphics[width=0.30\textwidth]{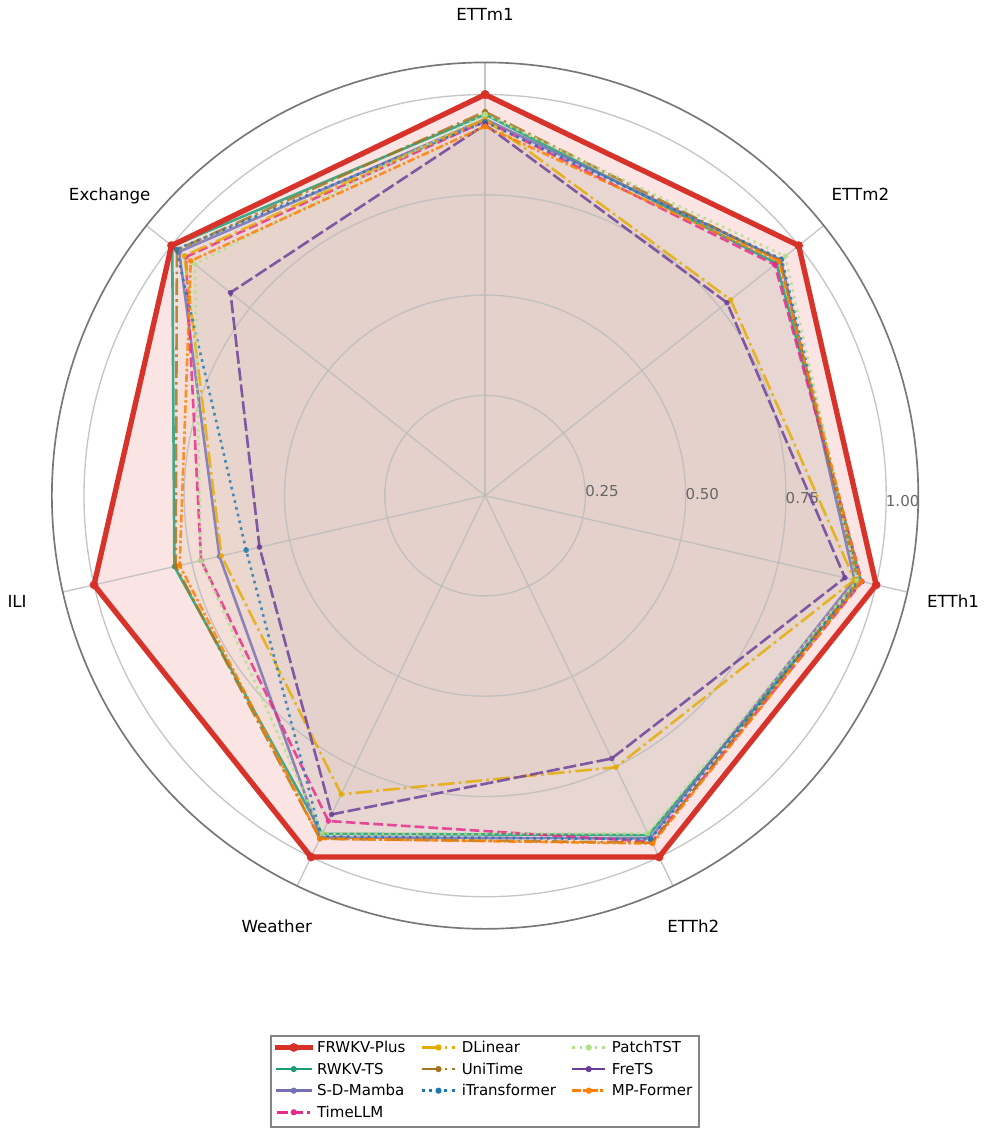}
\par\smallskip
\refstepcounter{figure}\label{fig:main_average_radar}
\small Figure~\thefigure: Radar view of Table~\ref{tab:main_average_comparison}. Left: MSE; right: MAE. Larger radius indicates lower relative error.
\end{minipage}
\end{center}

\section{Related Work}
\textbf{Long-term time series forecasting.}
Long-term multivariate time series forecasting has been studied through several generations of methods. Recent surveys summarize the field as moving from statistical and recurrent models toward attention-based, decomposition-based, frequency-aware, and foundation-model-style systems \cite{wen2023transformers}. Classical methods remain important reference points: ARIMA-family models characterize linear temporal structure \cite{box2015time}, vector autoregression models multivariate linear interactions \cite{stock2001var}, and GARCH-style models describe volatility dynamics \cite{bollerslev1986garch}. Neural sequence models improve nonlinear representation capacity, with recurrent architectures such as LSTM designed to retain temporal state \cite{hochreiter1997lstm} and probabilistic autoregressive methods extending neural forecasting to uncertainty-aware prediction \cite{salinas2020deepar}. Self-attention then provided a flexible mechanism for long-range dependency modeling \cite{vaswani2017attention}, and time-series representation learning further adapted Transformer blocks to multivariate temporal data \cite{zerveas2021transformer}.

Transformer-based forecasters are a major direction because they can organize long temporal contexts through attention, decomposition, or tokenization strategies. To make attention more suitable for long sequences, subsequent studies reduce attention complexity, correct non-stationary behavior, or reorganize observations into more effective temporal units. Informer is a representative sparse-attention model \cite{zhou2021informer}, while Non-stationary Transformer explicitly studies distribution shift in forecasting \cite{liu2022nonstationary}. Crossformer segments time series across temporal and variable dimensions to model cross-dimension dependency \cite{zhang2023crossformer}, while iTransformer inverts the tokenization perspective to model variables as tokens \cite{liu2024itransformer}. TimeMixer shows that decomposable multiscale mixing can remain competitive without relying on a conventional Transformer stack \cite{wang2024timemixer}. These developments have improved the modeling capacity of long-term forecasting systems, but their primary focus remains temporal representation learning.

Beyond attention-centered models, recent forecasting research has explored lightweight and alternative backbones. Linear and MLP-based forecasters show that simple structures can remain competitive on standard long-term forecasting benchmarks when temporal patterns are appropriately represented \cite{zeng2023dlinear}. Patch-based modeling reduces the burden of long-horizon representation by grouping neighboring observations into local temporal units \cite{nie2023patchtst}. Recurrent-style and state-space models provide another efficiency-oriented direction by avoiding dense pairwise token interactions \cite{hou2024rwkvts}. S-D-Mamba further examines whether Mamba-style sequence modeling is effective for forecasting tasks \cite{wang2025smamba}. Multiscale patch structures also reduce the burden of long-horizon modeling, as represented by MultiPatchFormer \cite{naghashi2025multipatchformer}. Taken together, these approaches broaden the design space of forecasting models, but they do not directly address the specific problem considered in this paper: how periodic-position evidence should be used to regulate the interaction between real and imaginary components in frequency-space forecasting.

Large pretrained and language-model-based forecasters extend the modeling scope by reusing broad sequence representations or cross-domain semantic cues. TimeLLM studies forecasting by reprogramming large language models \cite{jin2024timellm}, while UniTime pursues language-empowered representations for cross-domain time series forecasting \cite{liu2024unitime}. TimeGPT explores foundation-model-style forecasting over time series \cite{garza2023timegpt}. Chronos casts forecasting as sequence modeling over discretized time-series values \cite{ansari2024chronos}, and Chronos-2 extends this line toward multivariate and covariate-informed zero-shot forecasting \cite{ansari2025chronos2}. Cross-modality alignment methods such as TimeCMA \cite{liu2024timecma} and tri-modal designs such as T3Time \cite{chowdhury2025t3time} further show the value of combining temporal, textual, and spectral cues. These models provide a complementary route to broad forecasting ability, whereas FRWKV-Plus focuses on a compact mechanism for reliability-controlled periodic modulation inside a frequency-space forecaster.

\textbf{Efficient and frequency-space forecasting.}
Frequency-domain modeling is naturally connected to long-term time series forecasting because spectral transformations expose periodic and long-range structures in compact forms. Existing frequency-aware methods incorporate spectral information into decomposition, representation learning, or frequency-domain transformation pipelines. Autoformer connects decomposition with autocorrelation-based temporal dependency modeling \cite{wu2021autoformer}, and FEDformer further uses frequency-enhanced decomposition to capture long-term structure efficiently \cite{zhou2022fedformer}. ETSformer uses exponential smoothing to connect interpretable temporal components with Transformer forecasting \cite{woo2022etsformer}, while MICN combines local and global context modeling for long-term series prediction \cite{wang2023micn}. FreTS shows that frequency-domain MLPs can serve as effective learners for time series forecasting \cite{yi2023frets}. More recent designs combine frequency processing with multiscale convolutions, spectral filtering, or wavelet denoising to handle noisy and heterogeneous temporal patterns. FAMC-Net uses frequency-domain parity correction and multiscale dilated convolution \cite{wang2023famcnet}. SDformer combines spectral filtering with dynamic attention \cite{zhou2024sdformer}, and CAWformer computes frequency-domain cross-variable correlations with wavelet denoising \cite{fan2025cawformer}. Meanwhile, lightweight forecasting backbones aim to preserve efficiency while maintaining sufficient representation ability. FRWKV follows this line by using low-cost frequency-domain state-update blocks for long-term forecasting \cite{yang2025frwkv}.

However, frequency-space forecasting also introduces a structural issue that is less emphasized in prior work. After transformation into the complex spectrum, a real-valued time series is represented by coupled real and imaginary components. These two components jointly describe magnitude- and phase-related information, but efficient implementations often process them through separate streams for computational simplicity. This separation can limit the exchange of complementary information between the two branches. From a spectral perspective, temporal alignment changes are reflected through phase behavior, and phase variation is represented by the coordinated relationship between the real and imaginary parts. Local distortions may further affect both magnitude- and phase-related responses. Therefore, treating the two components as isolated streams may weaken phase-sensitive coordination in the frequency representation. Existing frequency-aware forecasters mainly focus on spectral decomposition, dominant frequency extraction, denoising, or standalone frequency representation learning, while explicit real-imaginary branch interaction remains underexplored. This motivates the cross-branch gating design in FRWKV-Plus.

\textbf{Periodicity-aware modeling and adaptive modulation.}
Periodicity is a central property of many real-world time series. Existing forecasting models commonly exploit periodic information through seasonal-trend decomposition, time-feature embedding, multi-period representation, or phase-aware token organization \cite{wu2023timesnet}. Seasonal-trend representation learning also appears in contrastive methods such as COST \cite{woo2022cost}, while CoGLformer uses global-local context alignment to preserve coarse temporal patterns for forecasting \cite{ding2025coglformer}. PPTformer provides another perspective by combining decomposition and specialized modules to better handle long-term patterns and extreme values \cite{liu2025pptformer}. These approaches demonstrate that periodic and contextual structure can improve forecasting when it provides reliable evidence about future variations. Nevertheless, periodic information is not always equally informative across samples, variables, and prediction horizons. Treating periodic cues as ordinary input features or fixed representation components may force the model to absorb periodic evidence even when its relevance to the current forecasting state is weak.

Adaptive modulation provides a more selective way to use such information. Instead of injecting periodic-position context directly into the whole representation, a model can use it to adjust a specific decision pathway. In FRWKV-Plus, this pathway is the interaction between real and imaginary frequency branches. The proposed TGPC differs from architectural reformulations such as multi-period decomposition, cross-modality alignment, or phase-aware tokenization: it uses compact periodic-position context as signed residual correction evidence for cross-branch gates, and regulates this correction with adaptive trust scores. In this sense, periodic information is not treated as a mandatory and rigid representation update, but as a flexible trust-controlled prior for refining cross-branch frequency interaction. This moves the design from deterministic period injection toward conditional spectral modulation.

Overall, existing studies have advanced long-term forecasting through stronger temporal representation, more efficient backbones, frequency-domain modeling, periodicity-aware designs, and foundation-model-based forecasting. FRWKV-Plus builds on these developments but focuses on a more specific mechanism: reliability-controlled use of periodic-position context for real-imaginary frequency branch interaction. This positioning distinguishes the proposed model from methods that primarily rely on attention efficiency, temporal decomposition, direct periodic feature injection, large-scale pretraining, or standalone frequency representation learning.

\section{Method}
\subsection{Problem Definition and Overall Framework}
Given a multivariate historical window $X\in\mathbb{R}^{B\times L\times N}$, where $B$ is the batch size, $L$ is the input length, and $N$ is the number of variables, the goal of long-term forecasting is to predict the future horizon $\hat{Y}\in\mathbb{R}^{B\times H\times N}$. FRWKV-Plus follows an efficient frequency-space forecasting pipeline. The input sequence is first normalized by RevIN, lifted into an embedding space, processed in the frequency domain, and finally projected to the prediction horizon.

\begin{figure}[t]
\centering
\includegraphics[width=\textwidth]{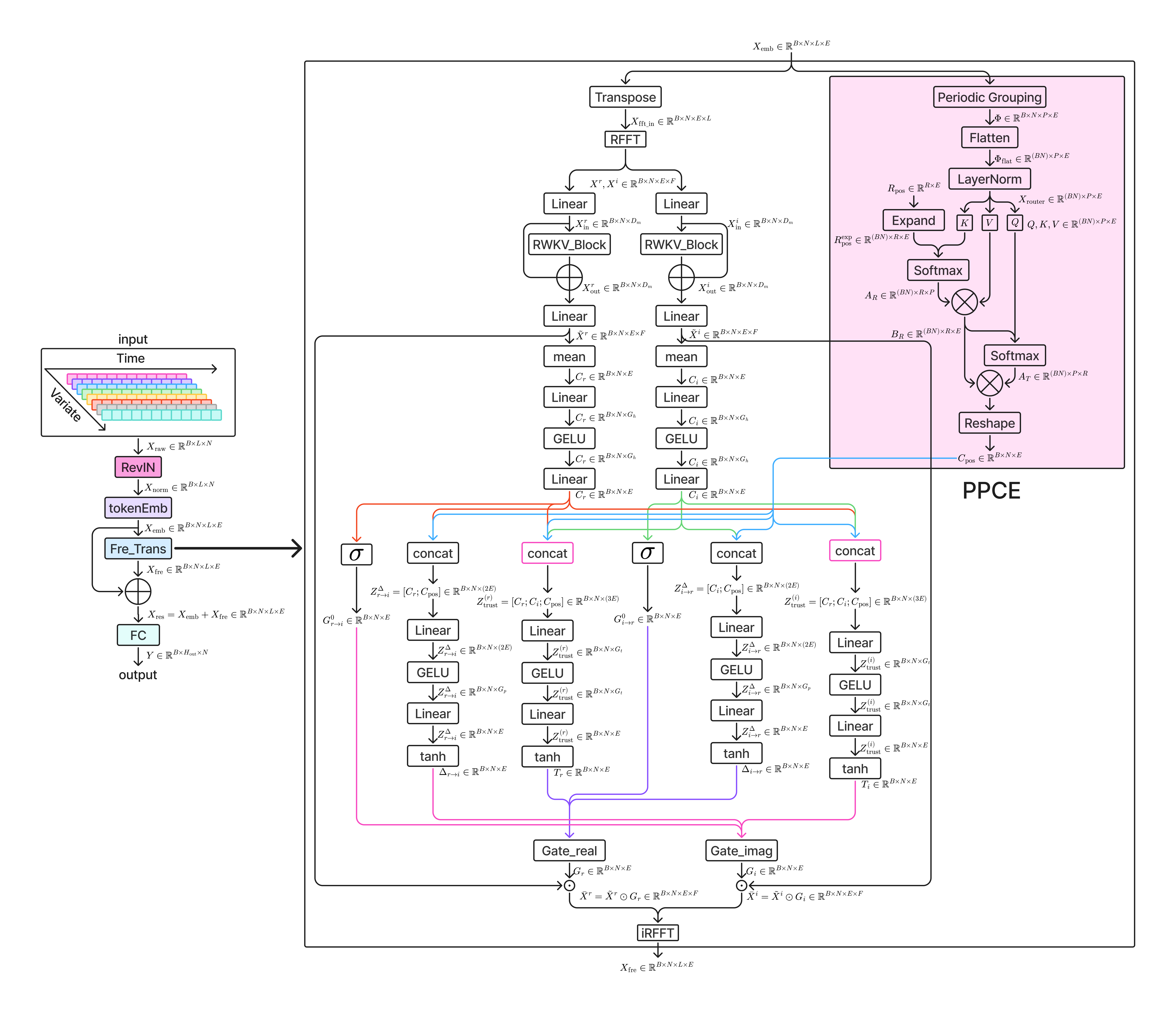}
\caption{Complete architecture of the FRWKV-Plus main model. The model applies RevIN and token embedding, sends the embedded sequence to the frequency transformation module, adds the transformed representation back to the embedding, and projects the result to the forecasting horizon. The frequency module contains real-imaginary RWKV encoding, cross-branch gating, PPCE-based periodic-position context extraction, signed gate correction, adaptive trust control, and inverse rFFT reconstruction.}
\label{fig:method_overview}
\end{figure}

Figure~\ref{fig:method_overview} summarizes the complete architecture. The overall design contains three key components. First, an inherited FRWKV frequency backbone encodes the real and imaginary parts of the complex spectrum with lightweight frequency branches. Second, cross-branch spectral gates let each branch condition the scaling of the other, introducing a lightweight coupling between the two spectral components. Third, the Periodic Positional Context Encoder (PPCE) and TGPC use compact periodic-position evidence to correct the cross-branch gates under adaptive trust control. The main methodological contribution is therefore not the reused frequency backbone itself, but the trust-controlled PPCE-to-gate correction pathway built on top of it.

\subsection{Frequency-Space FRWKV Backbone}
Before frequency transformation, FRWKV-Plus applies reversible instance normalization to reduce sample-wise and variable-wise distribution shifts \cite{kim2022revin}:
\begin{equation}
X_{\rm norm}[b,\ell,n]
=
\gamma_n
\frac{X[b,\ell,n]-\mu_{b,n}}{\sigma_{b,n}+\epsilon}
+\beta_n ,
\end{equation}
where $\mu_{b,n}$ and $\sigma_{b,n}$ are computed along the temporal dimension for each sample-variable pair, and $\gamma_n,\beta_n$ are learnable affine parameters. The normalized sequence is then lifted by a learnable embedding vector $a\in\mathbb{R}^{E}$:
\begin{equation}
X_{\rm emb}[b,n,\ell,e]
=
X_{\rm norm}[b,\ell,n]\,a_e,
\qquad
X_{\rm emb}\in\mathbb{R}^{B\times N\times L\times E}.
\end{equation}

The embedded sequence is first arranged as $X_{\rm emb}^{\rm T}\in\mathbb{R}^{B\times N\times E\times L}$ by swapping the embedding and temporal axes, and then transformed along the temporal axis by a normalized real FFT:
\begin{equation}
Z=\operatorname{rFFT}_{\ell}(X_{\rm emb}^{\rm T})
=
Z^{(r)}+iZ^{(i)},
\qquad
Z^{(r)},Z^{(i)}\in\mathbb{R}^{B\times N\times E\times F},
\end{equation}
where $\operatorname{rFFT}_{\ell}(\cdot)$ denotes real FFT along the temporal dimension with orthogonal normalization, and $F=\lfloor L/2\rfloor+1$ is the number of retained frequency points. The real and imaginary spectral tensors are encoded by two lightweight FRWKV frequency branches \cite{yang2025frwkv}. In the implementation used by FRWKV-Plus, each branch flattens the embedding-frequency dimensions, applies the corresponding FRWKV frequency encoder, adds a residual connection, and reshapes the output back to $\mathbb{R}^{B\times N\times E\times F}$:
\begin{equation}
\tilde{Z}^{(r)}
=
\operatorname{reshape}
\left(
\mathcal{F}_{r}(\operatorname{flat}(Z^{(r)}))
+
\operatorname{flat}(Z^{(r)})
\right),
\end{equation}
\begin{equation}
\tilde{Z}^{(i)}
=
\operatorname{reshape}
\left(
\mathcal{F}_{i}(\operatorname{flat}(Z^{(i)}))
+
\operatorname{flat}(Z^{(i)})
\right),
\end{equation}
where $\mathcal{F}_{r}$ and $\mathcal{F}_{i}$ denote the real and imaginary FRWKV frequency branches. These encoded responses are passed to the cross-branch gating modules below.

\subsection{Real-Imaginary Cross-Branch Gating}
The real and imaginary parts of a complex spectrum jointly encode magnitude- and phase-related information. Processing them independently is efficient, but it leaves each branch unaware of the state of the other. FRWKV-Plus therefore introduces lightweight cross-branch spectral gates (CSG), which let one branch condition the scaling of the other rather than exchange values directly. The encoded real and imaginary responses are first summarized along the frequency axis:
\begin{equation}
C_r=\operatorname{Mean}_{f}(\tilde{Z}^{(r)}),
\qquad
C_i=\operatorname{Mean}_{f}(\tilde{Z}^{(i)}),
\qquad
C_r,C_i\in\mathbb{R}^{B\times N\times E}.
\end{equation}

The base gates are generated from the opposite branch:
\begin{equation}
G_{i\rightarrow r}^{0}
=
\sigma\!\left(\operatorname{MLP}_{i\rightarrow r}^{0}(C_i)\right),
\qquad
G_{r\rightarrow i}^{0}
=
\sigma\!\left(\operatorname{MLP}_{r\rightarrow i}^{0}(C_r)\right).
\end{equation}
Here $G_{i\rightarrow r}^{0}$ is the base gate applied to the real branch using imaginary-branch context, and $G_{r\rightarrow i}^{0}$ is the base gate applied to the imaginary branch using real-branch context. The sigmoid restricts the base gate values to $(0,1)$, and the final scaling includes an identity path so that the gated response remains anchored to the original spectral encoding.

\subsection{Periodic Positional Context Encoder}
Periodic-position information is used as auxiliary evidence rather than as a direct representation update. Given a period-position length $P$, PPCE first groups the embedded sequence by within-period positions. If $L$ is not divisible by $P$, the sequence is padded by appending the first few time steps to the end, producing a length divisible by $P$. Let $M$ denote the resulting number of period groups. The repeated occurrences of the same within-period position are averaged:
\begin{equation}
\Phi[b,n,p,e]
=
\frac{1}{M}
\sum_{m=1}^{M}
X_{\rm pad}[b,n,m,p,e],
\qquad
\Phi\in\mathbb{R}^{B\times N\times P\times E}.
\end{equation}
Thus, $\Phi$ stores compact tokens for the $P$ within-period positions of each sample-variable pair.

To compress these tokens, PPCE uses a small learnable router set. After layer normalization and linear projections, the period-position tokens generate queries, keys, and values:
\begin{equation}
Q=X_{\rm pos}W_Q,\qquad
K=X_{\rm pos}W_K,\qquad
V=X_{\rm pos}W_V,
\end{equation}
where $X_{\rm pos}\in\mathbb{R}^{(BN)\times P\times E}$ is the reshaped and normalized form of $\Phi$. Let $R_{\rm pos}\in\mathbb{R}^{R\times E}$ be $R$ learnable router vectors. The routers first attend to period-position tokens:
\begin{equation}
B_R
=
\operatorname{softmax}
\left(
\frac{R_{\rm pos}K^\top}{\sqrt{E}}
\right)V,
\qquad
B_R\in\mathbb{R}^{(BN)\times R\times E}.
\end{equation}
The period-position queries then attend back to the router buffer:
\begin{equation}
U
=
\operatorname{softmax}
\left(
\frac{QB_R^\top}{\sqrt{E}}
\right)B_R.
\end{equation}
Finally, PPCE averages the router-enhanced period-position tokens and applies an output projection:
\begin{equation}
C_{\rm pos}
=
\operatorname{OutProj}
\left(
\operatorname{Mean}_{p}(U)
\right),
\qquad
C_{\rm pos}\in\mathbb{R}^{B\times N\times E}.
\end{equation}
The resulting $C_{\rm pos}$ is a compact periodic-position context, which we refer to as the within-period context (WPC). It has the same embedding width as the real and imaginary branch summaries, so it can be used to correct the cross-branch gates without creating an additional forecasting branch.

\subsection{Trust-Gated Periodic Correction}
TGPC converts the periodic-position context into signed residual corrections for the cross-branch gates. For the real-branch gate, the correction is generated from the imaginary-branch summary and periodic-position context; for the imaginary-branch gate, it is generated from the real-branch summary and periodic-position context:
\begin{equation}
\Delta_{i\rightarrow r}
=
\tanh
\left(
\operatorname{MLP}_{i\rightarrow r}^{\Delta}
([C_i,C_{\rm pos}])
\right),
\end{equation}
\begin{equation}
\Delta_{r\rightarrow i}
=
\tanh
\left(
\operatorname{MLP}_{r\rightarrow i}^{\Delta}
([C_r,C_{\rm pos}])
\right).
\end{equation}
The $\tanh$ activation makes the correction signed, allowing periodic evidence to either strengthen or weaken cross-branch exchange.

Because periodic cues are not uniformly reliable, FRWKV-Plus further learns branch-specific trust scores from the joint state of the real branch, imaginary branch, and periodic-position context:
\begin{equation}
T_r
=
\sigma
\left(
\operatorname{MLP}_{\rm trust}^{r}
([C_r,C_i,C_{\rm pos}])
\right),
\qquad
T_i
=
\sigma
\left(
\operatorname{MLP}_{\rm trust}^{i}
([C_r,C_i,C_{\rm pos}])
\right).
\end{equation}
These trust scores determine how strongly the signed corrections are admitted. The final gates are
\begin{equation}
G_r
=
1
+
G_{i\rightarrow r}^{0}
+
\alpha T_r\odot \Delta_{i\rightarrow r},
\end{equation}
\begin{equation}
G_i
=
1
+
G_{r\rightarrow i}^{0}
+
\alpha T_i\odot \Delta_{r\rightarrow i},
\end{equation}
where $\alpha$ is a learnable scalar clipped to $[0,0.20]$ during the forward pass. This bounded residual form preserves a stable base interaction path while allowing periodic-position context to adjust it selectively. Since $G^0\in(0,1)$, $T\in(0,1)$, $\Delta\in[-1,1]$, and $\alpha$ is small, the signed correction modifies the positive cross-branch scaling factor without turning the gate into an unconstrained negative modulation. In implementation, the final layers of the correction MLPs are initialized to zero, and the trust branches use a low initial bias, so FRWKV-Plus starts close to the base cross-branch interaction and gradually learns when periodic correction should be admitted.

The encoded spectral responses are then gated as
\begin{equation}
\bar{Z}^{(r)}
=
\tilde{Z}^{(r)}\odot G_r,
\qquad
\bar{Z}^{(i)}
=
\tilde{Z}^{(i)}\odot G_i,
\end{equation}
where $G_r$ and $G_i$ are broadcast only along the frequency axis while maintaining element-wise modulation across samples, variables, and embedding channels. The gated real and imaginary responses are recombined into a complex spectrum:
\begin{equation}
\bar{Z}
=
\bar{Z}^{(r)}+i\bar{Z}^{(i)}.
\end{equation}

\subsection{Forecast Reconstruction}
The corrected complex spectrum is mapped back to the time-domain embedding space by inverse rFFT with the same orthogonal normalization:
\begin{equation}
X_{\rm fre}
=
\left(\operatorname{irFFT}_{\ell}(\bar{Z})\right)^{\rm T}
\in\mathbb{R}^{B\times N\times L\times E}.
\end{equation}
Following the inherited FRWKV pipeline, the frequency-transformed representation is added back to the embedded input:
\begin{equation}
X_{\rm out}
=
X_{\rm emb}+X_{\rm fre}.
\end{equation}
A feed-forward prediction head then maps the flattened temporal-embedding representation to the forecasting horizon:
\begin{equation}
\hat{Y}
=
\operatorname{RevIN}^{-1}
\left(
\operatorname{Head}
(\operatorname{flat}(X_{\rm out}))
\right),
\qquad
\hat{Y}\in\mathbb{R}^{B\times H\times N}.
\end{equation}
This reconstruction step keeps the model aligned with the efficient frequency-space backbone, while the proposed TGPC modifies only the interaction between real and imaginary frequency branches. Consequently, FRWKV-Plus introduces periodic awareness through a lightweight correction mechanism rather than through a separate heavy representation path.

\clearpage

\section{Experiments}
This section evaluates FRWKV-Plus from three perspectives. First, we compare it with representative forecasting baselines on seven widely used long-term forecasting benchmarks to assess its overall predictive performance. Second, we conduct controlled module analyses to examine whether real-imaginary branch interaction, periodic-position context, signed gate correction, and adaptive trust control contribute to the final design. Third, we report efficiency profiles to verify whether the proposed periodic-aware gating mechanism preserves the lightweight behavior of the frequency-space backbone.

\subsection{Experimental Settings}
\paragraph{Datasets and forecasting protocol.}
We conduct experiments on seven standard multivariate time series forecasting benchmarks: ETTh1, ETTh2, ETTm1, ETTm2, Weather, Exchange, and ILI. These datasets cover electricity transformer temperature, meteorological observation, exchange-rate variation, and influenza-like illness records, and they are widely used in long-term forecasting studies. The forecasting protocol follows the common long-term forecasting setting adopted by decomposition-based forecasters \cite{wu2021autoformer}. For ETTh1, ETTh2, ETTm1, ETTm2, Weather, and Exchange, the input length is fixed to 96 and the prediction horizons are set to 96, 192, 336, and 720. For ILI, following the common illness forecasting setting, the input length is 36 and the prediction horizons are 24, 36, 48, and 60.

\paragraph{Evaluation metrics.}
We report mean squared error (MSE) and mean absolute error (MAE), which are standard metrics for long-term forecasting. MSE emphasizes larger forecasting deviations, while MAE reflects the average absolute deviation and is less sensitive to isolated large errors. Reporting both metrics provides a balanced view of accuracy and error stability across datasets and prediction horizons.

\paragraph{Compared methods.}
We compare FRWKV-Plus with representative methods from several forecasting directions, including recurrent-style and state-space forecasting models, large-model-based forecasting methods, lightweight linear models, frequency-domain forecasters, and Transformer-family architectures. This comparison is intended to evaluate whether the proposed periodic-aware frequency-space design remains competitive against both efficient forecasting backbones and more expressive sequence modeling architectures. All baseline values in the main comparison are reproduced locally under the unified forecasting protocol. To keep the comparison controlled, all models use the same input length (96, or 36 for ILI), prediction horizons, dataset splits, and evaluation metrics; we note that methods relying on substantially longer look-back windows or large-scale pre-training in their original settings may behave differently here, particularly on the small, domain-specific ILI dataset.

\paragraph{Implementation details.}
FRWKV-Plus is implemented in PyTorch and trained with AdamW using a cosine annealing learning-rate schedule. The learning rate is set to $1\times 10^{-4}$ and the weight decay is $1\times 10^{-3}$. The batch size is 32 for ETTh1, ETTh2, ETTm1, ETTm2, Weather, and Exchange, and 16 for ILI. The maximum training epochs and early-stopping patience are configured according to the forecasting setting. Early stopping is activated only after at least half of the scheduled epochs have elapsed, which avoids prematurely stopping before the frequency-space backbone and periodic-aware gates have reached stable adaptation. All experiments in this study were conducted on a server equipped with eight NVIDIA GeForce RTX 3090 GPUs.

For the reported FRWKV-Plus configurations, the hidden dimension and feed-forward dimension are both 512, the number of heads is 8, and the frequency embedding size is 16. The number of encoder layers is 2 for all datasets except Weather, for which it is 3. For the periodic-aware modules, unless otherwise stated we use a within-period length $P=24$, $R=4$ routers, a correction budget $\alpha$ initialized to $0.10$ and clipped to $[0,0.20]$ during the forward pass, and a trust bias initialized to $-2.0$; in addition, the final layers of the correction MLPs are zero-initialized, so the model starts at the identity-preserving, base-gate-only configuration and gradually learns when periodic correction should be admitted. The temporal patch length and stride follow the inherited FRWKV backbone. Training uses a weighted sequence loss (\texttt{loss\_mode=L1}, \texttt{lossfun\_alpha=0.5}) that places larger weights on earlier forecast steps while retaining MAE-style robustness; this loss is used only for optimizing FRWKV-Plus, and all models are evaluated with the standard MSE and MAE metrics.

\subsection{Overall Forecasting Performance}
Table~\ref{tab:main_average_comparison} reports the dataset-level average performance of FRWKV-Plus and representative baselines. For each dataset, the reported MSE/MAE values are obtained by equally averaging the four prediction horizons. This dataset-level summary gives a compact view of the overall forecasting behavior across the seven benchmarks, while the detailed horizon-level results for the same model set are provided in Tables~\ref{tab:detailed_horizon_results_state_lm} and~\ref{tab:detailed_horizon_results_linear_freq}.

The results show that FRWKV-Plus performs competitively across the seven benchmarks. Compared with lightweight linear and frequency-domain baselines, FRWKV-Plus improves the use of spectral information by introducing explicit real-imaginary branch interaction and periodic-aware gate correction. Compared with Transformer-family and large-model-based forecasters, FRWKV-Plus maintains a compact frequency-space design while achieving competitive or better average errors on multiple datasets. These results indicate that the proposed model provides a favorable balance between forecasting accuracy and structural efficiency.

The advantage of FRWKV-Plus is consistent with the motivation of the proposed design. Periodic-position information is not injected as an additional representation stream; instead, it is converted into a trust-controlled correction signal for the cross-branch gates. This allows the model to use recurring temporal evidence when it is compatible with the current spectral state, while preserving the stable frequency-space backbone. Therefore, the main comparison supports the central claim that selective periodic-aware modulation can strengthen frequency-space forecasting without introducing a heavy representation path.

\begin{table}[t]
\centering
\caption{Main comparison summarized by dataset-level average MSE/MAE.}
\label{tab:main_average_comparison}
\scriptsize
\setlength{\tabcolsep}{2.6pt}
\renewcommand{\arraystretch}{1.05}
\resizebox{\textwidth}{!}{%
\begin{tabular}{lccccccc}
\toprule
\multicolumn{1}{c}{\multirow{2}{*}{Model}} & \multicolumn{7}{c}{Dataset} \\
\cmidrule(lr){2-8}
& ETTm1 & ETTm2 & ETTh1 & ETTh2 & Weather & ILI & Exchange \\
\midrule
FRWKV-Plus & \textbf{\textcolor{red}{0.377}}/\textbf{\textcolor{red}{0.382}} & \textbf{\textcolor{red}{0.272}}/\textbf{\textcolor{red}{0.314}} & \textbf{\textcolor{red}{0.430}}/\textbf{\textcolor{red}{0.427}} & \textbf{\textcolor{red}{0.360}}/\textbf{\textcolor{red}{0.387}} & \textbf{\textcolor{red}{0.242}}/\textbf{\textcolor{red}{0.262}} & \textbf{\textcolor{red}{1.480}}/\textbf{\textcolor{red}{0.735}} & \textbf{\textcolor{red}{0.352}}/\textbf{\textcolor{red}{0.397}} \\
RWKV-TS~\cite{hou2024rwkvts} & 0.390/0.402 & 0.287/0.338 & 0.453/0.444 & \textbf{0.375}/0.412 & 0.256/0.280 & \textbf{1.910}/\textbf{0.925} & \textbf{0.353}/\textbf{0.398} \\
S-D-Mamba~\cite{wang2025smamba} & 0.398/0.407 & 0.290/0.333 & 0.457/0.452 & 0.383/0.408 & 0.252/0.277 & 2.664/1.082 & 0.364/0.407 \\
TimeLLM~\cite{jin2024timellm} & 0.410/0.409 & 0.296/0.340 & 0.448/\textbf{0.443} & 0.381/0.404 & 0.275/0.291 & 2.432/1.012 & 0.372/0.416 \\
DLinear~\cite{zeng2023dlinear} & 0.403/0.407 & 0.350/0.401 & 0.456/0.452 & 0.559/0.515 & 0.265/0.317 & 2.616/1.090 & 0.354/0.414 \\
UniTime~\cite{liu2024unitime} & \textbf{0.385}/\textbf{0.399} & 0.293/0.334 & \textbf{0.442}/0.448 & 0.378/0.403 & 0.253/\textbf{0.276} & 2.108/0.929 & 0.364/0.404 \\
iTransformer~\cite{liu2024itransformer} & 0.407/0.410 & 0.288/0.332 & 0.454/0.447 & 0.383/0.407 & 0.258/0.278 & 2.444/1.203 & 0.360/0.403 \\
PatchTST~\cite{nie2023patchtst} & 0.392/0.402 & \textbf{0.285}/\textbf{0.328} & 0.463/0.449 & 0.395/0.414 & 0.257/0.280 & 2.388/1.011 & 0.390/0.429 \\
FreTS~\cite{yi2023frets} & 0.406/0.413 & 0.375/0.407 & 0.476/0.464 & 0.584/0.532 & 0.255/0.297 & 3.424/1.274 & 0.509/0.489 \\
MP-Former~\cite{naghashi2025multipatchformer} & 0.413/0.415 & 0.291/0.335 & 0.446/0.444 & 0.378/\textbf{0.402} & \textbf{0.250}/\textbf{0.276} & 2.466/0.941 & 0.383/0.423 \\
\bottomrule
\end{tabular}}
\par\smallskip
\begin{minipage}{\textwidth}
\footnotesize \emph{Note:} Rows correspond to forecasting models and columns correspond to datasets. Each cell reports average MSE/MAE after equally averaging the four horizons of the corresponding dataset. MP-Former denotes MultiPatchFormer. Red marks the best value and bold black text marks the second-best value within the methods shown in this table; rounded ties may produce multiple marked entries.
\end{minipage}
\end{table}

\subsection{Ablation Study}
We quantify the contribution of each design choice through a single-variable ablation. All variants share the same training protocol and modify only one component relative to the full FRWKV-Plus design. Table~\ref{tab:ablation} reports dataset-level mean and standard deviation over three seeds. For each seed, MSE and MAE are first averaged over the four prediction horizons of the corresponding dataset, and the standard deviation is then computed across seeds. This setting allows us to distinguish consistent component effects from run-to-run variation.

The ablation rows are organized by mechanism type. B1 removes the base cross-branch spectral gate (CSG), B2 zeroes the within-period context (WPC), B3 replaces adaptive trust scores with a constant value, and B4 constrains signed correction to a positive-only form. C1 replaces the adaptive trust gate with an explicit usefulness estimator, while C2 replaces the single-period WPC with a multi-period context bank. D1, D2, and D3 examine channel-specific, linear-projection, and patch-fold embedding alternatives, respectively. These variants are used only to analyze the proposed design and are not additional models introduced by this paper.

\begin{table}[t]
\centering
\caption{Ablation study results with three-seed mean and standard deviation. Full is the proposed FRWKV-Plus model; all other rows are single-component ablations or design alternatives.}
\label{tab:ablation}
\scriptsize
\setlength{\tabcolsep}{1.8pt}
\renewcommand{\arraystretch}{1.04}
\resizebox{\textwidth}{!}{%
\begin{tabular}{llccccccc}
\toprule
\# & Variant & ETTm1 & ETTm2 & ETTh1 & ETTh2 & Weather & ILI & Exchange \\
\midrule
-- & \textbf{Full} & $\mathbf{0.377{\pm}0.002/0.382{\pm}0.002}$ & $\mathbf{0.272{\pm}0.002/0.314{\pm}0.001}$ & $\mathbf{0.430{\pm}0.003/0.427{\pm}0.001}$ & $\mathbf{0.360{\pm}0.003/0.387{\pm}0.002}$ & $\mathbf{0.242{\pm}0.000/0.262{\pm}0.000}$ & $\mathbf{1.480{\pm}0.010/0.735{\pm}0.003}$ & $\mathbf{0.352{\pm}0.017/0.397{\pm}0.009}$ \\
B1 & w/o CSG & $0.382{\pm}0.002/0.386{\pm}0.002$ & $0.275{\pm}0.002/0.316{\pm}0.001$ & $0.439{\pm}0.002/0.432{\pm}0.001$ & $0.370{\pm}0.006/0.391{\pm}0.003$ & $0.244{\pm}0.001/0.264{\pm}0.000$ & $1.625{\pm}0.024/0.769{\pm}0.007$ & $0.389{\pm}0.021/0.418{\pm}0.009$ \\
B2 & w/o WPC & $0.382{\pm}0.002/0.385{\pm}0.001$ & $0.274{\pm}0.001/0.316{\pm}0.001$ & $0.443{\pm}0.006/0.434{\pm}0.002$ & $0.369{\pm}0.005/0.391{\pm}0.002$ & $0.243{\pm}0.000/0.264{\pm}0.000$ & $1.635{\pm}0.031/0.772{\pm}0.008$ & $0.439{\pm}0.034/0.435{\pm}0.009$ \\
B3 & w/o Adaptive Trust & $0.382{\pm}0.003/0.385{\pm}0.002$ & $0.275{\pm}0.002/0.317{\pm}0.001$ & $0.443{\pm}0.004/0.434{\pm}0.001$ & $0.368{\pm}0.003/0.391{\pm}0.002$ & $0.243{\pm}0.001/0.264{\pm}0.001$ & $1.606{\pm}0.047/0.768{\pm}0.013$ & $0.396{\pm}0.006/0.420{\pm}0.004$ \\
B4 & Pos-only & $0.381{\pm}0.002/0.386{\pm}0.002$ & $0.276{\pm}0.001/0.317{\pm}0.000$ & $0.439{\pm}0.003/0.432{\pm}0.001$ & $0.369{\pm}0.000/0.391{\pm}0.001$ & $0.244{\pm}0.000/0.265{\pm}0.000$ & $1.624{\pm}0.024/0.773{\pm}0.006$ & $0.380{\pm}0.022/0.411{\pm}0.010$ \\
C1 & Explicit Reliability & $0.382{\pm}0.003/0.385{\pm}0.002$ & $0.276{\pm}0.000/0.317{\pm}0.000$ & $0.438{\pm}0.002/0.433{\pm}0.001$ & $0.367{\pm}0.002/0.389{\pm}0.001$ & $0.244{\pm}0.001/0.265{\pm}0.000$ & $1.636{\pm}0.002/0.777{\pm}0.002$ & $0.375{\pm}0.001/0.411{\pm}0.001$ \\
C2 & Multi-Period WPC & $0.380{\pm}0.000/0.384{\pm}0.000$ & $0.276{\pm}0.002/0.317{\pm}0.001$ & $0.440{\pm}0.001/0.433{\pm}0.001$ & $0.370{\pm}0.002/0.392{\pm}0.001$ & $0.244{\pm}0.001/0.264{\pm}0.000$ & $1.663{\pm}0.011/0.783{\pm}0.006$ & $0.391{\pm}0.021/0.417{\pm}0.008$ \\
D1 & Channel Emb. & $0.383{\pm}0.001/0.387{\pm}0.000$ & $0.277{\pm}0.001/0.317{\pm}0.000$ & $0.444{\pm}0.008/0.435{\pm}0.004$ & $0.368{\pm}0.004/0.392{\pm}0.001$ & $0.243{\pm}0.001/0.265{\pm}0.001$ & $1.642{\pm}0.008/0.770{\pm}0.001$ & $0.506{\pm}0.028/0.452{\pm}0.008$ \\
D2 & Linear Emb. & $0.382{\pm}0.001/0.385{\pm}0.001$ & $0.276{\pm}0.000/0.317{\pm}0.000$ & $0.439{\pm}0.003/0.433{\pm}0.003$ & $0.369{\pm}0.004/0.390{\pm}0.002$ & $0.244{\pm}0.002/0.265{\pm}0.001$ & $1.637{\pm}0.020/0.774{\pm}0.002$ & $0.426{\pm}0.048/0.429{\pm}0.018$ \\
D3 & Patch-Fold Emb. & $0.386{\pm}0.001/0.387{\pm}0.001$ & $0.276{\pm}0.001/0.317{\pm}0.001$ & $0.440{\pm}0.002/0.435{\pm}0.001$ & $0.370{\pm}0.003/0.391{\pm}0.002$ & $0.246{\pm}0.002/0.266{\pm}0.001$ & $1.671{\pm}0.050/0.784{\pm}0.014$ & $0.423{\pm}0.046/0.425{\pm}0.014$ \\
\bottomrule
\end{tabular}}
\par\smallskip
\begin{minipage}{\textwidth}
\footnotesize \emph{Note:} Each cell reports MSE/MAE. Values are dataset-level mean$\pm$std over three seeds. For each seed, the metric is first averaged over the four prediction horizons of the corresponding dataset. CSG denotes cross-branch spectral gating, WPC denotes within-period context, and Pos-only denotes positive-only correction.
\end{minipage}
\end{table}

\textbf{Overall contribution of the complete design.}
Across all seven datasets, FRWKV-Plus attains the lowest mean MSE and MAE. On the strongly periodic benchmarks its lead over the closest variant is small, within one to two standard deviations, whereas on Exchange and ILI the margin is large and clearly exceeds the three-seed variance. We therefore treat the easy-dataset ordering as indicative and base our component claims primarily on the harder datasets, where the effects are statistically clear. Overall, the proposed modules behave as mutually complementary refinements rather than a single dominant component.

\textbf{Effect size and stability across datasets.}
On the ETT subsets and Weather, the absolute differences among ablation variants are relatively small, often around one to three percent in MSE. The three-seed standard deviations show that fine ordering among variants on these benchmarks should be interpreted cautiously; for example, ETTm1 variants range from $0.380$ to $0.386$ in MSE, with standard deviations around $0.001$--$0.003$, compared with $0.377{\pm}0.002$ for the full model. The pattern is clearer on ILI and Exchange. On ILI, the strongest degradation reaches $1.671{\pm}0.050$, compared with $1.480{\pm}0.010$ for FRWKV-Plus. On Exchange, the degradation can reach $0.506{\pm}0.028$, compared with $0.352{\pm}0.017$. These larger gaps suggest that the proposed gating pathway is especially useful when the forecasting setting is more variable or data-limited. Importantly, these components add negligible cost, about $0.01$M parameters and essentially unchanged peak memory over the backbone (Table~\ref{tab:eff}), so their near-neutral effect on the easy datasets carries little downside, while their pronounced benefit on Exchange and ILI yields a favorable cost--benefit trade-off.

\textbf{Role of periodic-position context and cross-branch gating.}
Among the removal variants, zeroing the within-period context (B2) yields one of the largest average degradations, most visibly on Exchange, where MSE increases from $0.352{\pm}0.017$ to $0.439{\pm}0.034$. Removing the cross-branch spectral gate (B1) is also damaging, reaching $0.389{\pm}0.021$ on the same dataset. Because Exchange shows larger run-to-run variation, we avoid over-interpreting the fine ordering among B1--B4. Nevertheless, the results consistently show that both WPC and CSG carry substantial information for the final gating behavior. The prominence of WPC even on the weakly periodic Exchange is, at first sight, counter-intuitive; we hypothesize that the cross-period averaged context acts less as a literal seasonal pattern than as a low-variance summary that stabilizes the cross-branch correction, and we leave a mechanistic verification of this effect to future work.

\textbf{Trust, correction sign, and alternative designs.}
Fixing the trust score (B3) or restricting the correction to a positive-only form (B4) also degrades the full model, although the penalties are generally smaller than removing the periodic context or cross-branch gate. This supports the use of signed, trust-modulated correction as a refinement mechanism. Replacing the adaptive trust gate with an explicit reliability predictor (C1) gives a stable alternative, with small standard deviations on ILI and Exchange, but it does not surpass the full model. The multi-period context variant (C2) is also consistently worse than FRWKV-Plus, indicating that the compact single-period context is sufficient for the present design and that additional period scales may introduce unnecessary noise.

\textbf{Embedding alternatives.}
The embedding substitutions in D1--D3 do not improve the complete design. Channel-specific, linear-projection, and patch-fold embeddings all produce higher errors than the inherited lightweight embedding, with particularly clear degradation on Exchange. This result supports retaining the shared lightweight embedding of the frequency-space backbone, which provides a robust interface for the periodic-position correction pathway.

In summary, the ablation study supports a consistent interpretation: FRWKV-Plus combines several lightweight components whose effects are modest but stable on regular periodic benchmarks and more pronounced on challenging datasets. The complete design occupies a favorable accuracy--robustness operating point because neither richer correction contexts, explicit reliability replacements, nor alternative embeddings improve upon the proposed configuration.

\subsection{Efficiency Analysis}
Table~\ref{tab:eff} compares the parameter count and peak training memory of FRWKV, FRWKV+CSG (the backbone with only the cross-branch spectral gate), and FRWKV-Plus on representative settings. FRWKV-Plus preserves the lightweight parameter scale of the FRWKV backbone: the parameter increase is about 0.01M relative to FRWKV in the reported settings, and peak training memory remains nearly unchanged.

The intended trade-off is visible in the measurement: FRWKV-Plus adds the TGPC correction module while keeping the memory profile close to the frequency-space linear backbone. Its empirical benefit should be interpreted together with the module analysis, where the small parameter and memory overhead is connected to adaptive periodic-position correction.

\begin{table}[t]
\centering
\caption{Parameter and memory usage on representative settings.}
\label{tab:eff}
\scriptsize
\begin{tabular}{llcc}
\toprule
Model & Dataset & Params(M) & Peak train MB \\
\midrule
FRWKV & ETTh1-96 & 14.357 & 486.1 \\
FRWKV+CSG & ETTh1-96 & 14.359 & 486.1 \\
FRWKV-Plus & ETTh1-96 & 14.369 & 487.5 \\
FRWKV & ETTh2-192 & 14.381 & 486.7 \\
FRWKV+CSG & ETTh2-192 & 14.384 & 486.8 \\
FRWKV-Plus & ETTh2-192 & 14.393 & 488.0 \\
\bottomrule
\end{tabular}
\par\smallskip
\begin{minipage}{\textwidth}
\footnotesize \emph{Note:} Params(M) reports trainable parameter count in millions. Peak train MB reports peak allocated CUDA memory during local training under the same environment.
\end{minipage}
\end{table}

\subsection{Qualitative Forecasting Analysis}
\begin{figure}[t]
\centering
\includegraphics[width=0.95\textwidth]{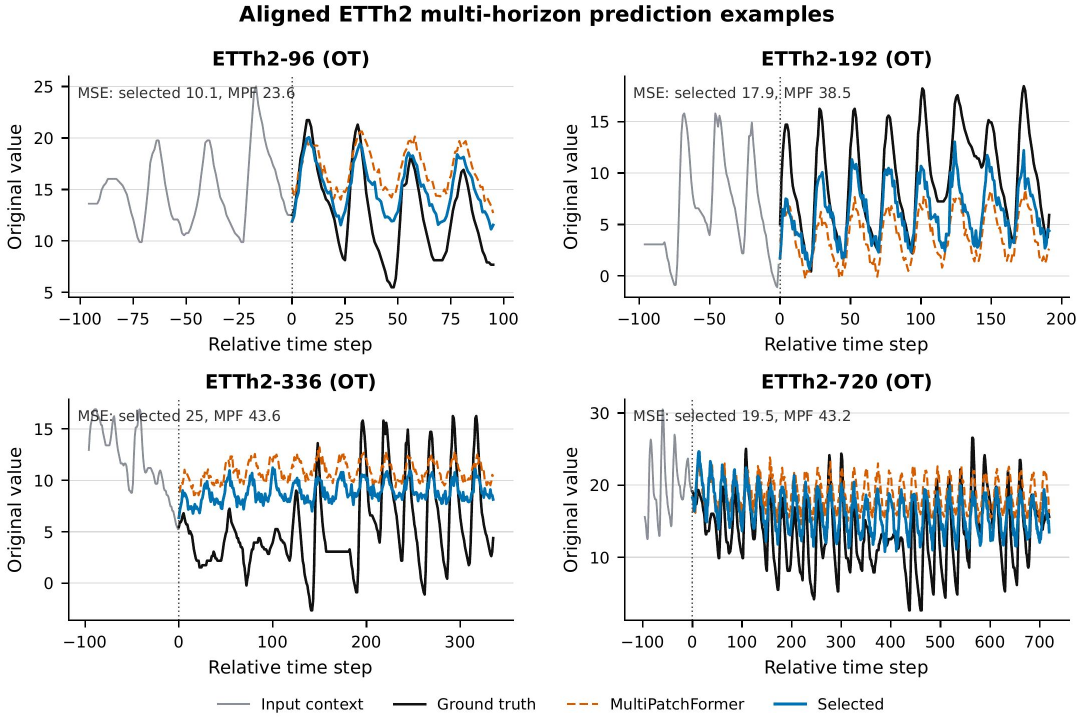}
\caption{Aligned multi-horizon prediction examples. Each panel compares the input context, ground truth, FRWKV-Plus prediction, and MultiPatchFormer prediction on the same target channel and aligned test window. Window-level MSE values in the panels are computed on the original data scale for the displayed target channel only; they are qualitative window diagnostics and are not directly comparable across panels.}
\label{fig:prediction_examples}
\end{figure}

To complement the table-level metrics, Figure~\ref{fig:prediction_examples} visualizes aligned future prediction windows from a representative benchmark across the four standard horizons. The curves compare FRWKV-Plus with MultiPatchFormer on the same target channel and aligned test windows after mapping predictions to the original data scale.

The visualization is used as qualitative evidence rather than as a statistical test. It illustrates how the proposed frequency-space design tracks future trajectory changes in representative windows, while the aggregate benchmark tables remain the basis for the quantitative performance claims.

\subsection{Discussion}
The ablation gives a more precise interpretation of FRWKV-Plus than the benchmark tables alone. The controlled comparisons indicate that the gain does not come from a single dominant module, but from several lightweight, mutually complementary components, among which the within-period context is the most influential. They also show that the benefit is modest on strongly periodic data and concentrated on the more challenging Exchange and ILI settings, supporting the view that periodic-position context is most useful as a trust-controlled, bounded correction rather than as an unconditional representation update.

The results also clarify why the correction is implemented as an adaptive gate rather than as unconditional feature injection. Periodic-position summaries are useful because they provide compact evidence about recurring temporal structure, while signed corrections allow the model to adjust cross-branch gates in both directions. Adaptive trust further controls the correction strength, connecting periodic evidence with the frequency-space representation in a lightweight and explicit manner.

This use pattern is important for reliability-aware prediction systems because it connects a structured temporal cue with an explicit admission mechanism. FRWKV-Plus does not require periodic-position context to dominate the representation. Instead, it uses that context to correct frequency-branch gates when the evidence is informative, while preserving the efficiency of the FRWKV backbone. Future extensions can make periodic-position reliability more explicit and extend the evaluation to larger and more heterogeneous forecasting datasets.

\paragraph{Limitations.}
This study is intentionally scoped to seven standard long-term forecasting benchmarks: ETTh1, ETTh2, ETTm1, ETTm2, Weather, Exchange, and ILI. These datasets cover several commonly used forecasting regimes, but they do not exhaust the behavior of large-scale, high-dimensional, or strongly heterogeneous industrial time series. In particular, Traffic-like datasets and larger high-dimensional benchmarks remain important future test cases because they can stress cross-variable interaction, memory use, and the reliability of compact periodic-position summaries more severely than the datasets studied here.

Another limitation is that the current adaptive trust mechanism estimates the reliability of periodic-position evidence implicitly through learned gates rather than through an explicit reliability estimator. Future work can make this reliability estimation more transparent, study larger high-dimensional datasets, and further improve the balance between compact periodic correction and cross-variable context.

Finally, several design and evaluation choices are intentionally kept fixed rather than tuned. The within-period length is fixed to $P=24$ for all datasets for consistency, even though the natural period differs across sampling granularities; the multi-period variant (C2 in Table~\ref{tab:ablation}) did not improve over this single-period setting, but a granularity-aware choice of $P$ remains an open direction. In addition, all methods are compared under a fixed input length, and the main comparison reports protocol-level averages; the three-seed variance is characterized in the ablation study, while a fuller variance analysis for every baseline under multiple look-back windows is left to future work.

\section{Conclusion}
This work studied reliability-controlled periodic-position correction for frequency-space linear time series forecasting. FRWKV-Plus keeps the efficient FRWKV backbone, adds explicit real-imaginary branch interaction, and uses TGPC to derive signed gate corrections from periodic-position context under adaptive trust control. Benchmark comparisons show that FRWKV-Plus is consistently competitive across seven standard datasets, while three-seed ablations show that each component contributes, that the within-period context is the most influential single component, and that the benefit concentrates on the more challenging datasets. These findings support the view that periodic-position information is valuable when used as a selective, trust-controlled correction signal. Future work should develop more principled estimators of periodic-position reliability and extend the analysis to larger and more heterogeneous forecasting datasets.

\section*{CRediT authorship contribution statement}
Qingyuan Yang: Conceptualization, Methodology, Validation, Formal analysis, Writing - original draft, Writing - review \& editing.
Dongyue Chen: Supervision, Project administration, Writing - review \& editing.
Da Teng: Validation, Writing - review \& editing.
Junhua Xiao: Validation, Writing - review \& editing.
Jiaji Pan: Validation, Writing - review \& editing.
Shizhuo Deng: Supervision, Writing - review \& editing.

\section*{Funding}
This work was supported by the National Key R\&D Program of China (2024YFB4710900) and the Guangdong Basic and Applied Basic Research Foundation under Grant 2024A1515010244.

\section*{Declaration of competing interest}
The authors declare that they have no known competing financial interests or personal relationships that could have appeared to influence the work reported in this paper.

\section*{Data availability}
The benchmark datasets used in this study are publicly available long-term forecasting datasets. The code, configuration files, and result aggregation scripts supporting this study are publicly available at \url{https://github.com/yangqingyuan-byte/FRWKV-plus}.

\section*{Declaration of generative AI use}
During the preparation of this work, the authors used generative artificial intelligence tools to assist with coding tasks related to experiment organization and result-file processing. The authors reviewed and verified the AI-assisted coding outputs and take full responsibility for the content of the published article.

\clearpage
\appendix
\section{Detailed Horizon-Level Forecasting Results}
Tables~\ref{tab:detailed_horizon_results_state_lm} and~\ref{tab:detailed_horizon_results_linear_freq} report the horizon-level MSE and MAE values corresponding to the model set summarized in Table~\ref{tab:main_average_comparison}. The average rows are obtained by equally averaging the four prediction horizons of each dataset.

\begin{table}[p]
\centering
\caption{Detailed horizon-level forecasting results for FRWKV-Plus, recurrent-style, state-space, large-model, and linear baselines.}
\label{tab:detailed_horizon_results_state_lm}
\scriptsize
\setlength{\tabcolsep}{1.8pt}
\renewcommand{\arraystretch}{1.02}
\resizebox{\textwidth}{!}{%
\begin{tabular}{c|c|cc|cc|cc|cc|cc}
\toprule
\multirow{2}{*}{\textbf{Dataset}} & \multirow{2}{*}{\textbf{Horizon}} & \multicolumn{2}{c|}{\shortstack{\textbf{FRWKV-Plus}}} & \multicolumn{2}{c|}{\shortstack{\textbf{RWKV-TS}\\\textbf{(2024)}}} & \multicolumn{2}{c|}{\shortstack{\textbf{S-D-Mamba}\\\textbf{(2024)}}} & \multicolumn{2}{c|}{\shortstack{\textbf{TimeLLM}\\\textbf{(2024)}}} & \multicolumn{2}{c}{\shortstack{\textbf{DLinear}\\\textbf{(2023)}}} \\
& & MSE & MAE & MSE & MAE & MSE & MAE & MSE & MAE & MSE & MAE \\
\midrule
\multirow{5}{*}{ETTm1}
& 96 & \textbf{\textcolor{red}{0.308}} & \textbf{\textcolor{red}{0.339}} & 0.327 & 0.365 & 0.331 & 0.368 & 0.359 & 0.381 & 0.345 & 0.372 \\
& 192 & \textbf{\textcolor{red}{0.357}} & \textbf{\textcolor{red}{0.370}} & 0.369 & 0.388 & 0.378 & 0.393 & 0.383 & 0.393 & 0.380 & 0.389 \\
& 336 & \textbf{\textcolor{red}{0.389}} & \textbf{\textcolor{red}{0.391}} & 0.402 & 0.409 & 0.410 & 0.414 & 0.416 & 0.414 & 0.413 & 0.413 \\
& 720 & \textbf{\textcolor{red}{0.452}} & \textbf{\textcolor{red}{0.427}} & 0.461 & 0.446 & 0.474 & 0.451 & 0.483 & 0.449 & 0.474 & 0.453 \\
& Avg & \textbf{\textcolor{red}{0.377}} & \textbf{\textcolor{red}{0.382}} & 0.390 & 0.402 & 0.398 & 0.407 & 0.410 & 0.409 & 0.403 & 0.407 \\
\midrule
\multirow{5}{*}{ETTm2}
& 96 & \textbf{\textcolor{red}{0.171}} & \textbf{\textcolor{red}{0.249}} & 0.178 & 0.270 & 0.182 & 0.266 & 0.193 & 0.280 & 0.193 & 0.292 \\
& 192 & \textbf{\textcolor{red}{0.233}} & \textbf{\textcolor{red}{0.291}} & 0.251 & 0.314 & 0.252 & 0.313 & 0.257 & 0.318 & 0.284 & 0.362 \\
& 336 & \textbf{0.292} & \textbf{0.328} & 0.310 & 0.353 & 0.313 & 0.349 & 0.317 & 0.353 & 0.369 & 0.427 \\
& 720 & \textbf{\textcolor{red}{0.390}} & \textbf{\textcolor{red}{0.388}} & \textbf{0.407} & 0.414 & 0.413 & \textbf{0.405} & 0.419 & 0.411 & 0.554 & 0.522 \\
& Avg & \textbf{\textcolor{red}{0.272}} & \textbf{\textcolor{red}{0.314}} & 0.287 & 0.338 & 0.290 & 0.333 & 0.296 & 0.340 & 0.350 & 0.401 \\
\midrule
\multirow{5}{*}{ETTh1}
& 96 & \textbf{\textcolor{red}{0.371}} & \textbf{\textcolor{red}{0.388}} & \textbf{0.384} & \textbf{0.400} & 0.388 & 0.406 & 0.398 & 0.410 & 0.386 & \textbf{0.400} \\
& 192 & \textbf{\textcolor{red}{0.424}} & \textbf{\textcolor{red}{0.419}} & 0.445 & 0.433 & 0.445 & 0.441 & 0.451 & 0.440 & 0.437 & 0.432 \\
& 336 & \textbf{\textcolor{red}{0.462}} & \textbf{\textcolor{red}{0.441}} & 0.488 & 0.459 & 0.490 & 0.465 & 0.473 & \textbf{0.451} & 0.481 & 0.459 \\
& 720 & \textbf{\textcolor{red}{0.463}} & \textbf{\textcolor{red}{0.460}} & 0.496 & 0.484 & 0.506 & 0.497 & \textbf{0.469} & \textbf{0.470} & 0.519 & 0.516 \\
& Avg & \textbf{\textcolor{red}{0.430}} & \textbf{\textcolor{red}{0.427}} & 0.453 & 0.444 & 0.457 & 0.452 & 0.448 & \textbf{0.443} & 0.456 & 0.452 \\
\midrule
\multirow{5}{*}{ETTh2}
& 96 & \textbf{\textcolor{red}{0.278}} & \textbf{\textcolor{red}{0.327}} & 0.311 & 0.364 & 0.297 & 0.349 & 0.295 & 0.345 & 0.333 & 0.387 \\
& 192 & \textbf{\textcolor{red}{0.358}} & \textbf{\textcolor{red}{0.379}} & 0.376 & 0.410 & 0.378 & 0.399 & 0.386 & 0.399 & 0.477 & 0.476 \\
& 336 & \textbf{0.396} & \textbf{\textcolor{red}{0.410}} & \textbf{\textcolor{red}{0.390}} & \textbf{0.420} & 0.425 & 0.435 & 0.419 & 0.429 & 0.594 & 0.541 \\
& 720 & \textbf{\textcolor{red}{0.409}} & \textbf{\textcolor{red}{0.430}} & \textbf{0.421} & 0.454 & 0.432 & 0.448 & 0.425 & \textbf{0.442} & 0.831 & 0.657 \\
& Avg & \textbf{\textcolor{red}{0.360}} & \textbf{\textcolor{red}{0.387}} & \textbf{0.375} & 0.412 & 0.383 & 0.408 & 0.381 & 0.404 & 0.559 & 0.515 \\
\midrule
\multirow{5}{*}{Weather}
& 96 & \textbf{\textcolor{red}{0.156}} & \textbf{\textcolor{red}{0.194}} & 0.178 & 0.221 & \textbf{0.165} & \textbf{0.209} & 0.198 & 0.235 & 0.196 & 0.255 \\
& 192 & \textbf{\textcolor{red}{0.205}} & \textbf{\textcolor{red}{0.238}} & 0.219 & 0.256 & 0.215 & 0.255 & 0.240 & 0.269 & 0.237 & 0.296 \\
& 336 & \textbf{\textcolor{red}{0.264}} & \textbf{\textcolor{red}{0.282}} & 0.275 & 0.297 & 0.273 & 0.296 & 0.295 & 0.308 & 0.283 & 0.335 \\
& 720 & \textbf{\textcolor{red}{0.342}} & \textbf{\textcolor{red}{0.335}} & 0.353 & 0.347 & 0.353 & 0.349 & 0.368 & 0.353 & \textbf{0.345} & 0.381 \\
& Avg & \textbf{\textcolor{red}{0.242}} & \textbf{\textcolor{red}{0.262}} & 0.256 & 0.280 & 0.252 & 0.277 & 0.275 & 0.291 & 0.265 & 0.317 \\
\midrule
\multirow{5}{*}{ILI}
& 24 & \textbf{\textcolor{red}{1.432}} & \textbf{\textcolor{red}{0.721}} & \textbf{2.036} & \textbf{0.916} & 2.675 & 1.074 & 2.383 & 1.004 & 2.398 & 1.040 \\
& 36 & \textbf{\textcolor{red}{1.392}} & \textbf{\textcolor{red}{0.714}} & \textbf{1.916} & 0.920 & 2.578 & 1.059 & 2.390 & 0.993 & 2.646 & 1.088 \\
& 48 & \textbf{\textcolor{red}{1.467}} & \textbf{\textcolor{red}{0.730}} & \textbf{1.896} & 0.937 & 2.668 & 1.084 & 2.394 & 1.003 & 2.614 & 1.086 \\
& 60 & \textbf{\textcolor{red}{1.628}} & \textbf{\textcolor{red}{0.773}} & \textbf{1.790} & 0.927 & 2.735 & 1.110 & 2.562 & 1.049 & 2.804 & 1.146 \\
& Avg & \textbf{\textcolor{red}{1.480}} & \textbf{\textcolor{red}{0.735}} & \textbf{1.910} & \textbf{0.925} & 2.664 & 1.082 & 2.432 & 1.012 & 2.616 & 1.090 \\
\midrule
\multirow{5}{*}{Exchange}
& 96 & \textbf{\textcolor{red}{0.081}} & \textbf{\textcolor{red}{0.198}} & \textbf{0.084} & \textbf{0.202} & 0.086 & 0.206 & 0.087 & 0.208 & 0.088 & 0.218 \\
& 192 & \textbf{\textcolor{red}{0.172}} & \textbf{\textcolor{red}{0.296}} & 0.180 & 0.300 & 0.182 & 0.304 & \textbf{0.173} & \textbf{0.299} & 0.176 & 0.315 \\
& 336 & \textbf{0.315} & \textbf{\textcolor{red}{0.405}} & 0.338 & 0.421 & 0.331 & 0.417 & 0.375 & 0.454 & \textbf{\textcolor{red}{0.313}} & 0.427 \\
& 720 & 0.840 & \textbf{0.690} & \textbf{\textcolor{red}{0.811}} & \textbf{\textcolor{red}{0.671}} & 0.858 & 0.699 & 0.853 & 0.703 & \textbf{0.839} & 0.695 \\
& Avg & \textbf{\textcolor{red}{0.352}} & \textbf{\textcolor{red}{0.397}} & \textbf{0.353} & \textbf{0.398} & 0.364 & 0.407 & 0.372 & 0.416 & 0.354 & 0.414 \\
\bottomrule
\end{tabular}}
\par\smallskip
\begin{minipage}{\textwidth}
\footnotesize \emph{Note:} Each cell reports MSE and MAE for a specific prediction horizon. Avg rows report the equally weighted average across the four horizons of the corresponding dataset. Red marks the best value and bold black text marks the second-best value within the full Table~\ref{tab:main_average_comparison} model set; rounded ties may produce multiple marked entries.
\end{minipage}
\end{table}

\begin{table}[p]
\centering
\caption{Detailed horizon-level forecasting results for FRWKV-Plus, language-empowered, Transformer-family, frequency-domain, and multiscale baselines.}
\label{tab:detailed_horizon_results_linear_freq}
\scriptsize
\setlength{\tabcolsep}{1.55pt}
\renewcommand{\arraystretch}{1.02}
\resizebox{\textwidth}{!}{%
\begin{tabular}{c|c|cc|cc|cc|cc|cc|cc}
\toprule
\multirow{2}{*}{\textbf{Dataset}} & \multirow{2}{*}{\textbf{Horizon}} & \multicolumn{2}{c|}{\shortstack{\textbf{FRWKV-Plus}}} & \multicolumn{2}{c|}{\shortstack{\textbf{UniTime}\\\textbf{(2024)}}} & \multicolumn{2}{c|}{\shortstack{\textbf{iTransformer}\\\textbf{(2024)}}} & \multicolumn{2}{c|}{\shortstack{\textbf{PatchTST}\\\textbf{(2023)}}} & \multicolumn{2}{c|}{\shortstack{\textbf{FreTS}\\\textbf{(2023)}}} & \multicolumn{2}{c}{\shortstack{\textbf{MP-Former}\\\textbf{(2025)}}} \\
& & MSE & MAE & MSE & MAE & MSE & MAE & MSE & MAE & MSE & MAE & MSE & MAE \\
\midrule
\multirow{5}{*}{ETTm1}
& 96 & \textbf{\textcolor{red}{0.308}} & \textbf{\textcolor{red}{0.339}} & \textbf{0.322} & \textbf{0.363} & 0.334 & 0.368 & 0.344 & 0.373 & 0.339 & 0.374 & 0.345 & 0.377 \\
& 192 & \textbf{\textcolor{red}{0.357}} & \textbf{\textcolor{red}{0.370}} & \textbf{0.366} & 0.387 & 0.377 & 0.391 & 0.367 & \textbf{0.386} & 0.383 & 0.398 & 0.384 & 0.395 \\
& 336 & \textbf{\textcolor{red}{0.389}} & \textbf{\textcolor{red}{0.391}} & 0.398 & \textbf{0.407} & 0.426 & 0.420 & \textbf{0.392} & \textbf{0.407} & 0.419 & 0.422 & 0.427 & 0.425 \\
& 720 & \textbf{\textcolor{red}{0.452}} & \textbf{\textcolor{red}{0.427}} & \textbf{0.454} & \textbf{0.440} & 0.491 & 0.459 & 0.464 & 0.442 & 0.484 & 0.460 & 0.494 & 0.463 \\
& Avg & \textbf{\textcolor{red}{0.377}} & \textbf{\textcolor{red}{0.382}} & \textbf{0.385} & \textbf{0.399} & 0.407 & 0.410 & 0.392 & 0.402 & 0.406 & 0.413 & 0.413 & 0.415 \\
\midrule
\multirow{5}{*}{ETTm2}
& 96 & \textbf{\textcolor{red}{0.171}} & \textbf{\textcolor{red}{0.249}} & 0.183 & 0.266 & 0.180 & 0.264 & \textbf{0.177} & \textbf{0.260} & 0.196 & 0.287 & 0.180 & 0.264 \\
& 192 & \textbf{\textcolor{red}{0.233}} & \textbf{\textcolor{red}{0.291}} & 0.251 & 0.310 & 0.250 & 0.309 & \textbf{0.246} & \textbf{0.305} & 0.329 & 0.392 & 0.248 & 0.308 \\
& 336 & \textbf{0.292} & \textbf{0.328} & \textbf{\textcolor{red}{0.183}} & \textbf{\textcolor{red}{0.266}} & 0.311 & 0.348 & 0.305 & 0.343 & 0.405 & 0.427 & 0.315 & 0.353 \\
& 720 & \textbf{\textcolor{red}{0.390}} & \textbf{\textcolor{red}{0.388}} & 0.420 & 0.410 & 0.412 & 0.407 & 0.410 & \textbf{0.405} & 0.570 & 0.524 & 0.421 & 0.413 \\
& Avg & \textbf{\textcolor{red}{0.272}} & \textbf{\textcolor{red}{0.314}} & 0.293 & 0.334 & 0.288 & 0.332 & \textbf{0.285} & \textbf{0.328} & 0.375 & 0.407 & 0.291 & 0.335 \\
\midrule
\multirow{5}{*}{ETTh1}
& 96 & \textbf{\textcolor{red}{0.371}} & \textbf{\textcolor{red}{0.388}} & 0.397 & 0.418 & 0.386 & 0.405 & 0.404 & 0.413 & 0.396 & 0.408 & 0.391 & 0.410 \\
& 192 & \textbf{\textcolor{red}{0.424}} & \textbf{\textcolor{red}{0.419}} & \textbf{0.434} & 0.439 & 0.441 & 0.436 & 0.454 & \textbf{0.430} & 0.451 & 0.443 & 0.436 & 0.436 \\
& 336 & \textbf{\textcolor{red}{0.462}} & \textbf{\textcolor{red}{0.441}} & \textbf{0.468} & 0.457 & 0.487 & 0.458 & 0.497 & 0.462 & 0.501 & 0.473 & 0.485 & 0.461 \\
& 720 & \textbf{\textcolor{red}{0.463}} & \textbf{\textcolor{red}{0.460}} & \textbf{0.469} & 0.477 & 0.503 & 0.491 & 0.496 & 0.481 & 0.555 & 0.532 & 0.473 & \textbf{0.470} \\
& Avg & \textbf{\textcolor{red}{0.430}} & \textbf{\textcolor{red}{0.427}} & \textbf{0.442} & 0.448 & 0.454 & 0.447 & 0.463 & 0.449 & 0.476 & 0.464 & 0.446 & 0.444 \\
\midrule
\multirow{5}{*}{ETTh2}
& 96 & \textbf{\textcolor{red}{0.278}} & \textbf{\textcolor{red}{0.327}} & 0.296 & 0.345 & 0.297 & 0.349 & 0.312 & 0.358 & 0.372 & 0.417 & \textbf{0.294} & \textbf{0.342} \\
& 192 & \textbf{\textcolor{red}{0.358}} & \textbf{\textcolor{red}{0.379}} & 0.374 & 0.394 & 0.380 & 0.400 & 0.397 & 0.408 & 0.528 & 0.506 & \textbf{0.373} & \textbf{0.392} \\
& 336 & \textbf{0.396} & \textbf{\textcolor{red}{0.410}} & 0.415 & 0.427 & 0.428 & 0.432 & 0.435 & 0.440 & 0.570 & 0.531 & 0.414 & 0.424 \\
& 720 & \textbf{\textcolor{red}{0.409}} & \textbf{\textcolor{red}{0.430}} & 0.425 & 0.444 & 0.427 & 0.445 & 0.436 & 0.449 & 0.865 & 0.676 & 0.431 & 0.448 \\
& Avg & \textbf{\textcolor{red}{0.360}} & \textbf{\textcolor{red}{0.387}} & 0.378 & 0.403 & 0.383 & 0.407 & 0.395 & 0.414 & 0.584 & 0.532 & 0.378 & \textbf{0.402} \\
\midrule
\multirow{5}{*}{Weather}
& 96 & \textbf{\textcolor{red}{0.156}} & \textbf{\textcolor{red}{0.194}} & 0.171 & 0.214 & 0.174 & 0.214 & 0.177 & 0.218 & 0.185 & 0.239 & 0.167 & \textbf{0.209} \\
& 192 & \textbf{\textcolor{red}{0.205}} & \textbf{\textcolor{red}{0.238}} & 0.217 & \textbf{0.254} & 0.221 & \textbf{0.254} & 0.222 & 0.259 & 0.223 & 0.274 & \textbf{0.214} & \textbf{0.254} \\
& 336 & \textbf{\textcolor{red}{0.264}} & \textbf{\textcolor{red}{0.282}} & 0.274 & \textbf{0.293} & 0.278 & 0.296 & 0.277 & 0.297 & \textbf{0.270} & 0.312 & 0.271 & 0.296 \\
& 720 & \textbf{\textcolor{red}{0.342}} & \textbf{\textcolor{red}{0.335}} & 0.351 & \textbf{0.343} & 0.358 & 0.349 & 0.352 & 0.347 & \textbf{\textcolor{red}{0.342}} & 0.364 & 0.347 & 0.346 \\
& Avg & \textbf{\textcolor{red}{0.242}} & \textbf{\textcolor{red}{0.262}} & 0.253 & \textbf{0.276} & 0.258 & 0.278 & 0.257 & 0.280 & 0.255 & 0.297 & \textbf{0.250} & \textbf{0.276} \\
\midrule
\multirow{5}{*}{ILI}
& 24 & \textbf{\textcolor{red}{1.432}} & \textbf{\textcolor{red}{0.721}} & 2.346 & 0.954 & 2.347 & 1.731 & 2.335 & 0.989 & 3.220 & 1.229 & 3.226 & 1.034 \\
& 36 & \textbf{\textcolor{red}{1.392}} & \textbf{\textcolor{red}{0.714}} & 1.998 & \textbf{0.912} & 2.468 & 0.998 & 2.561 & 1.035 & 3.314 & 1.249 & 2.582 & 0.968 \\
& 48 & \textbf{\textcolor{red}{1.467}} & \textbf{\textcolor{red}{0.730}} & 1.979 & 0.912 & 2.489 & 1.016 & 2.465 & 1.022 & 3.425 & 1.277 & 1.963 & \textbf{0.852} \\
& 60 & \textbf{\textcolor{red}{1.628}} & \textbf{\textcolor{red}{0.773}} & 2.109 & 0.938 & 2.471 & 1.065 & 2.189 & 0.997 & 3.739 & 1.339 & 2.092 & \textbf{0.909} \\
& Avg & \textbf{\textcolor{red}{1.480}} & \textbf{\textcolor{red}{0.735}} & 2.108 & 0.929 & 2.444 & 1.203 & 2.388 & 1.011 & 3.424 & 1.274 & 2.466 & 0.941 \\
\midrule
\multirow{5}{*}{Exchange}
& 96 & \textbf{\textcolor{red}{0.081}} & \textbf{\textcolor{red}{0.198}} & 0.086 & 0.209 & 0.086 & 0.206 & 0.109 & 0.236 & 0.094 & 0.223 & 0.098 & 0.225 \\
& 192 & \textbf{\textcolor{red}{0.172}} & \textbf{\textcolor{red}{0.296}} & 0.174 & \textbf{0.299} & 0.177 & \textbf{0.299} & 0.205 & 0.327 & 0.235 & 0.357 & 0.195 & 0.320 \\
& 336 & \textbf{0.315} & \textbf{\textcolor{red}{0.405}} & 0.319 & \textbf{0.408} & 0.331 & 0.417 & 0.356 & 0.436 & 0.599 & 0.588 & 0.350 & 0.430 \\
& 720 & 0.840 & \textbf{0.690} & 0.875 & 0.701 & 0.847 & 0.691 & 0.888 & 0.716 & 1.107 & 0.790 & 0.889 & 0.716 \\
& Avg & \textbf{\textcolor{red}{0.352}} & \textbf{\textcolor{red}{0.397}} & 0.364 & 0.404 & 0.360 & 0.403 & 0.390 & 0.429 & 0.509 & 0.489 & 0.383 & 0.423 \\
\bottomrule
\end{tabular}}
\par\smallskip
\begin{minipage}{\textwidth}
\footnotesize \emph{Note:} Each cell reports MSE and MAE for a specific prediction horizon. Avg rows report the equally weighted average across the four horizons of the corresponding dataset. Red marks the best value and bold black text marks the second-best value within the full Table~\ref{tab:main_average_comparison} model set; rounded ties may produce multiple marked entries.
\end{minipage}
\end{table}


\begin{thebibliography}{99}
\bibitem{gao2023adaptive} Gao J, Chen Y, Hu W, Zhang D. An adaptive deep-learning load forecasting framework by integrating transformer and domain knowledge. Advances in Applied Energy. 2023;10:100142.
\bibitem{ji2023spatiotemporal} Ji J, Wang J, Huang C, Wu J, Xu B, Wu Z, Zhang J, Zheng Y. Spatiotemporal self-supervised learning for traffic flow prediction. Proceedings of the AAAI Conference on Artificial Intelligence. 2023;37:4356--4364.
\bibitem{cheng2022financial} Cheng D, Yang F, Xiang S, Liu J. Financial time series forecasting with multimodality graph neural network. Pattern Recognition. 2022;121:108218.
\bibitem{wu2023interpretableweather} Wu H, Zhou H, Long M, Wang J. Interpretable weather forecasting for worldwide stations with a unified deep model. Nature Machine Intelligence. 2023;5(6):602--611.
\bibitem{morid2023healthcare} Morid MA, Sheng ORL, Dunbar J. Time series prediction using deep learning methods in healthcare. ACM Transactions on Management Information Systems. 2023;14(1):1--29.
\bibitem{wen2023transformers} Wen Q, Zhou T, Zhang C, Chen W, Ma Z, Yan J, Sun L. Transformers in time series: a survey. In: Proceedings of the International Joint Conference on Artificial Intelligence; 2023. p. 6778--6786.
\bibitem{yang2025frwkv} Yang Q, Deng S, Chen D, Teng D, Gan Z. FRWKV: frequency-domain linear attention for long-term time series forecasting. arXiv preprint arXiv:2512.07539; 2025. doi:10.48550/arXiv.2512.07539. Available at: \url{https://arxiv.org/abs/2512.07539}.
\bibitem{box2015time} Box GEP, Jenkins GM, Reinsel GC, Ljung GM. Time Series Analysis: Forecasting and Control. 5th ed. Hoboken: Wiley; 2015.
\bibitem{stock2001var} Stock JH, Watson MW. Vector autoregressions. Journal of Economic Perspectives. 2001;15(4):101--115.
\bibitem{bollerslev1986garch} Bollerslev T. Generalized autoregressive conditional heteroskedasticity. Journal of Econometrics. 1986;31(3):307--327.
\bibitem{hochreiter1997lstm} Hochreiter S, Schmidhuber J. Long short-term memory. Neural Computation. 1997;9(8):1735--1780.
\bibitem{salinas2020deepar} Salinas D, Flunkert V, Gasthaus J, Januschowski T. DeepAR: probabilistic forecasting with autoregressive recurrent networks. International Journal of Forecasting. 2020;36(3):1181--1191.
\bibitem{vaswani2017attention} Vaswani A, Shazeer N, Parmar N, Uszkoreit J, Jones L, Gomez AN, Kaiser L, Polosukhin I. Attention is all you need. In: Advances in Neural Information Processing Systems; 2017.
\bibitem{zerveas2021transformer} Zerveas G, Jayaraman S, Patel D, Bhamidipaty A, Eickhoff C. A transformer-based framework for multivariate time series representation learning. In: Proceedings of the 27th ACM SIGKDD Conference on Knowledge Discovery and Data Mining; 2021. p. 2114--2124.
\bibitem{zhou2021informer} Zhou H, Zhang S, Peng J, Zhang S, Li J, Xiong H, Zhang W. Informer: beyond efficient Transformer for long sequence time-series forecasting. Proceedings of the AAAI Conference on Artificial Intelligence. 2021;35(12):11106--11115.
\bibitem{liu2022nonstationary} Liu Y, Wu H, Wang J, Long M. Non-stationary Transformers: exploring the stationarity in time series forecasting. Advances in Neural Information Processing Systems. 2022;35:9881--9893.
\bibitem{zhang2023crossformer} Zhang Y, Yan J. Crossformer: Transformer utilizing cross-dimension dependency for multivariate time series forecasting. In: International Conference on Learning Representations; 2023.
\bibitem{liu2024itransformer} Liu Y, Hu T, Zhang H, Wu H, Wang S, Ma L, Long M. iTransformer: inverted transformers are effective for time series forecasting. In: International Conference on Learning Representations; 2024.
\bibitem{wang2024timemixer} Wang S, Wu H, Shi X, Hu T, Luo H, Ma L, Zhang JY, Zhou J. TimeMixer: decomposable multiscale mixing for time series forecasting. In: International Conference on Learning Representations; 2024.
\bibitem{zeng2023dlinear} Zeng A, Chen M, Zhang L, Xu Q. Are transformers effective for time series forecasting? Proceedings of the AAAI Conference on Artificial Intelligence. 2023;37(9):11121--11128.
\bibitem{nie2023patchtst} Nie Y, Nguyen NH, Sinthong P, Kalagnanam J. A time series is worth 64 words: long-term forecasting with transformers. In: International Conference on Learning Representations; 2023.
\bibitem{hou2024rwkvts} Hou H, Yu FR. RWKV-TS: beyond traditional recurrent neural network for time series tasks. arXiv preprint arXiv:2401.09093; 2024. doi:10.48550/arXiv.2401.09093. Available at: \url{https://arxiv.org/abs/2401.09093}.
\bibitem{wang2025smamba} Wang Z, Kong F, Feng S, Wang M, Yang X, Zhao H, Wang D, Zhang Y. Is Mamba effective for time series forecasting? Neurocomputing. 2025;619:129178.
\bibitem{naghashi2025multipatchformer} Naghashi V, Boukadoum M, Diallo AB. A multiscale model for multivariate time series forecasting. Scientific Reports. 2025;15:1565.
\bibitem{jin2024timellm} Jin M, Wang S, Ma L, Chu Z, Zhang JY, Shi X, Chen P-Y, Liang Y, Li Y-F, Pan S, Wen Q. Time-LLM: time series forecasting by reprogramming large language models. In: International Conference on Learning Representations; 2024.
\bibitem{liu2024unitime} Liu X, Hu J, Li Y, Diao S, Liang Y, Hooi B, Zimmermann R. UniTime: a language-empowered unified model for cross-domain time series forecasting. In: Proceedings of the ACM Web Conference; 2024. p. 4095--4106.
\bibitem{garza2023timegpt} Garza A, Challu C, Mergenthaler-Canseco M. TimeGPT-1. arXiv preprint arXiv:2310.03589; 2023.
\bibitem{ansari2024chronos} Ansari AF, Stella L, Turkmen C, Zhang X, Mercado P, Shen H, et al. Chronos: learning the language of time series. arXiv preprint arXiv:2403.07815; 2024.
\bibitem{ansari2025chronos2} Ansari AF, Shchur O, Kuken J, Auer A, Han B, Mercado P, et al. Chronos-2: from univariate to universal forecasting. arXiv preprint arXiv:2510.15821; 2025. doi:10.48550/arXiv.2510.15821.
\bibitem{liu2024timecma} Liu C, Xu Q, Miao H, Yang S, Zhang L, Long C, Li Z, Zhao R. TimeCMA: towards LLM-empowered multivariate time series forecasting via cross-modality alignment. Proceedings of the AAAI Conference on Artificial Intelligence. 2025;39(18):18780--18788. doi:10.1609/aaai.v39i18.34067.
\bibitem{chowdhury2025t3time} Chowdhury AM, Akter R, Arib SH. T3Time: tri-modal time series forecasting via adaptive multi-head alignment and residual fusion. Proceedings of the AAAI Conference on Artificial Intelligence. 2026;40(25):20597--20605. doi:10.1609/aaai.v40i25.39196.
\bibitem{wu2021autoformer} Wu H, Xu J, Wang J, Long M. Autoformer: decomposition Transformers with auto-correlation for long-term series forecasting. Advances in Neural Information Processing Systems. 2021;34:22419--22430.
\bibitem{zhou2022fedformer} Zhou T, Ma Z, Wen Q, Wang X, Sun L, Jin R. FEDformer: frequency enhanced decomposed Transformer for long-term series forecasting. In: Proceedings of the 39th International Conference on Machine Learning; 2022. p. 27268--27286.
\bibitem{woo2022etsformer} Woo G, Liu C, Sahoo D, Kumar A, Hoi S. ETSformer: exponential smoothing Transformers for time-series forecasting. arXiv preprint arXiv:2202.01381; 2022.
\bibitem{wang2023micn} Wang H, Peng J, Huang F, Wang J, Chen J, Xiao Y. MICN: multi-scale local and global context modeling for long-term series forecasting. In: International Conference on Learning Representations; 2023.
\bibitem{yi2023frets} Yi K, Zhang Q, Fan W, Wang S, Wang P, He H, An N, Lian D, Cao L, Niu Z. Frequency-domain MLPs are more effective learners in time series forecasting. Advances in Neural Information Processing Systems. 2023;36:76656--76679.
\bibitem{wang2023famcnet} Wang M, Wang H, Zhang F. FAMC-Net: frequency domain parity correction attention and multi-scale dilated convolution for time series forecasting. In: Proceedings of the 32nd ACM International Conference on Information and Knowledge Management; 2023. p. 2554--2563.
\bibitem{zhou2024sdformer} Zhou Z, Lyu G, Huang Y, Wang Z, Jia Z, Yang Z. SDformer: Transformer with spectral filter and dynamic attention for multivariate time series long-term forecasting. In: Proceedings of the Thirty-Third International Joint Conference on Artificial Intelligence; 2024. p. 3--9.
\bibitem{fan2025cawformer} Fan S, Wang H, Zhang F. CAWformer: a cross variable attention with discrete wavelet denoising for multivariate time series forecasting. Knowledge-Based Systems. 2025;324:113846. doi:10.1016/j.knosys.2025.113846.
\bibitem{wu2023timesnet} Wu H, Hu T, Liu Y, Zhou H, Wang J, Long M. TimesNet: temporal 2D-variation modeling for general time series analysis. In: International Conference on Learning Representations; 2023.
\bibitem{woo2022cost} Woo G, Liu C, Sahoo D, Kumar A, Hoi S. COST: contrastive learning of disentangled seasonal-trend representations for time series forecasting. arXiv preprint arXiv:2202.01575; 2022.
\bibitem{ding2025coglformer} Ding F, Xu C, Liu H, Lyu C, Yang G, Xiong H, Zhou H. Global-local coherency contrastive learning for context-aware time series forecasting. Knowledge-Based Systems. 2026;331:114745. doi:10.1016/j.knosys.2025.114745.
\bibitem{liu2025pptformer} Liu J, Guo J, Gao L, Wang Y, Liu A, Zhang X. PPTformer: a novel hybrid model for enhanced long-term time series forecasting with extreme value focus. Knowledge-Based Systems. 2025;317:113456. doi:10.1016/j.knosys.2025.113456.
\bibitem{kim2022revin} Kim T, Kim J, Tae Y, Park C, Choi J, Choo J. Reversible instance normalization for accurate time-series forecasting against distribution shift. In: International Conference on Learning Representations; 2022.
\end{thebibliography}
\end{document}